\documentclass[10pt,twocolumn,letterpaper]{article}

\usepackage[pagenumbers]{wacv} % To force page numbers, e.g. for an arXiv version

\usepackage{graphicx}
\usepackage{amsmath}
\usepackage{amssymb}
\usepackage{booktabs}
\usepackage{pifont}
\usepackage{tabularx}
\usepackage{graphbox}
\PassOptionsToPackage{svgnames}{xcolor}
\usepackage{colortbl}
\usepackage{multirow}
\usepackage{subcaption}
\usepackage{comment}
\usepackage{svg}
\usepackage{mathtools}

\usepackage[pagebackref,breaklinks,colorlinks]{hyperref}

\usepackage[capitalize]{cleveref}
\crefname{section}{Sec.}{Secs.}
\Crefname{section}{Section}{Sections}
\Crefname{table}{Table}{Tables}
\crefname{table}{Tab.}{Tabs.}

\def\wacvPaperID{2657} % *** Enter the WACV Paper ID here
\def\confName{WACV}
\def\confYear{2025}

\newcommand{\cmark}{\ding{51}}%
\newcommand{\xmark}{\ding{55}}%

\newcommand{\benchmarkacron}{PrACo}
\newcommand{\benchmarkname}{Prompt-Aware Counting}

\begin{document}

\title{Mind the Prompt: A Novel Benchmark for Prompt-based Class-Agnostic Counting}

% \author{Luca Ciampi\thanks{Corresponding authors. They contribute equally to this work.}\\
% ISTI-CNR, Pisa, Italy\\
% %Institution1 address\\
% {\tt\small luca.ciampi@isti.cnr.it}
% \and
% Nicola Messina$^\ast$\\
% ISTI-CNR, Pisa, Italy\\
% %First line of institution2 address\\
% {\tt\small nicola.messina@isti.cnr.it}
% \and
% Matteo Pierucci\\
% University of Pisa, Italy\\
% %First line of institution2 address\\
% {\tt\small m.pierucci5@studenti.unipi.it}
% \and
% Giuseppe Amato\\
% ISTI-CNR, Pisa, Italy\\
% %First line of institution2 address\\
% {\tt\small giuseppe.amato@isti.cnr.it}
% \and
% Marco Avvenuti\\
% University of Pisa, Italy\\
% %First line of institution2 address\\
% {\tt\small marco.avvenuti@unipi.it}
% \and
% Fabrizio Falchi\\
% ISTI-CNR, Pisa, Italy\\
% %First line of institution2 address\\
% {\tt\small fabrizio.falchi@isti.cnr.it}
% }

% \author{
% \textbf{Luca Ciampi$^{1 \ast}$, \qquad Nicola Messina$^{1 \ast}$, \qquad Matteo Pierucci$^2$} \\  \textbf{Giuseppe Amato$^1$, \qquad Marco Avvenuti$^2$, \qquad Fabrizio Falchi$^1$} \\
% $^1$CNR-ISTI, Pisa, Italy \quad $^2$University of Pisa, Italy\\
% %\tt\small luca.ciampi@isti.cnr.it, nicola.messina@isti.cnr.it, m.pierucci5@studenti.unipi.it \\
% %\tt\small giuseppe.amato@isti.cnr.it, marco.avvenuti@unipi.it, fabrizio.falchi@isti.cnr.it
% }%

\author{
\begin{tabular}{>{\centering\arraybackslash}p{4cm}>{\centering\arraybackslash}p{4cm}>{\centering\arraybackslash}p{4cm}}
\textbf{Luca Ciampi$^{1 \ast}$} & \textbf{Nicola Messina$^{1 \ast}$} & \textbf{Matteo Pierucci$^2$} \\
\textbf{Giuseppe Amato$^1$} & \textbf{Marco Avvenuti$^2$} & \textbf{Fabrizio Falchi$^1$} \\
\end{tabular} \\
$^1$CNR-ISTI, Pisa, Italy \quad $^2$University of Pisa, Italy \\
\tt\small luca.ciampi@isti.cnr.it, nicola.messina@isti.cnr.it
}

\maketitle
\def\thefootnote{*}\footnotetext{Corresponding authors, they contributed equally to this work.} %\\ \tt\scriptsize luca.ciampi@isti.cnr.it, nicola.messina@isti.cnr.it}\def\thefootnote{\arabic{footnote}}

%%%%%%%%%%%%%%%%%%%%%%%%%%%% ABSTRACT %%%%%%%%%%%%%%%%%%%%%%%%%%%%%%%%%%%%%%%%%%%
\begin{abstract}
Recently, object counting has shifted towards class-agnostic counting (CAC), which counts instances of arbitrary object classes never seen during model training. With advancements in robust vision-and-language foundation models, there is a growing interest in \textit{prompt-based} CAC, where object categories are specified using natural language.
However, we identify significant limitations in current benchmarks for evaluating this task, which hinder both accurate assessment and the development of more effective solutions. Specifically, we argue that the current evaluation protocols do not measure the ability of the model to understand \textit{which} object has to be counted. This is due to two main factors: (i) the shortcomings of CAC datasets, which primarily consist of images containing objects from a single class, and (ii) the limitations of current counting performance evaluators, which are based on traditional class-specific counting and focus solely on counting errors.
To fill this gap, we introduce the \benchmarkname{} (\benchmarkacron{}) benchmark. It comprises two targeted tests coupled with evaluation metrics specifically designed to quantitatively measure the robustness and trustworthiness of existing prompt-based CAC models. We evaluate state-of-the-art methods and demonstrate that, although some achieve impressive results on standard class-specific counting metrics, they exhibit a significant deficiency in understanding the input prompt, indicating the need for more careful training procedures or revised designs.
The code for reproducing our results is available at \url{https://github.com/ciampluca/PrACo}.
\end{abstract}

\begin{figure}[t]
    \centering
    \includegraphics[width=.95\linewidth,page=2]{imgs/images.pdf}
    \includegraphics[width=\linewidth]{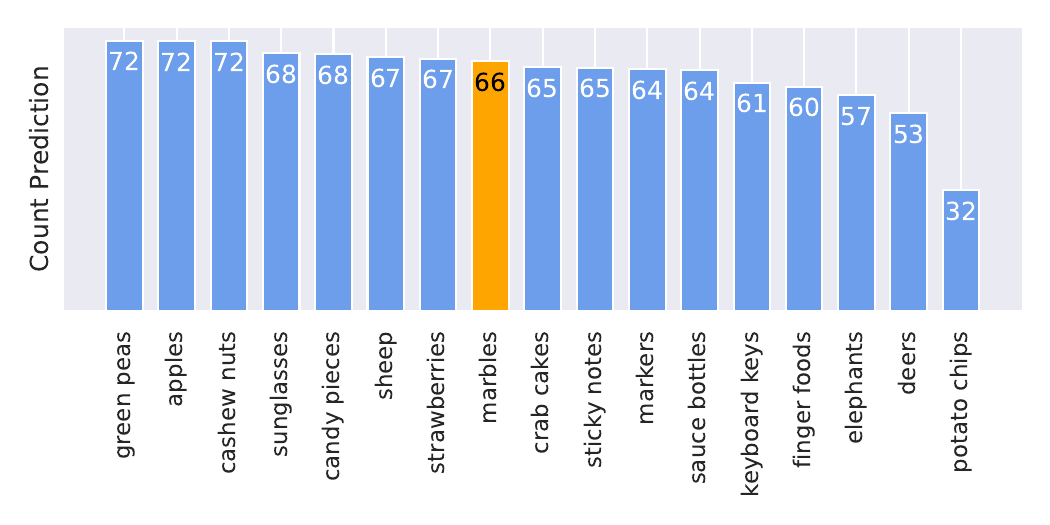}
    \caption{Prompt-based counting models -- CounTX \cite{AminiNaieni23} in this example -- exhibit difficulties in accurately interpreting user-provided texts that specify object classes to be counted. The confusion occurs even between classes that are semantically very distinct -- like \textit{marbles} and \textit{elephants}. In some cases, the count of classes not present in the image is even higher than that for the ground-truth object category (highlighted in orange).}
    \label{fig:double-inference}
\end{figure}

%%%%%%%%%%%%%%%%%%%%%%%%%%%% INTRODUCTION %%%%%%%%%%%%%%%%%%%%%%%%%%%%%%%%%%%%%%%%%%%
\section{Introduction}
\label{sec:intro}

Class-agnostic counting (CAC) aims to count instances of arbitrary objects beyond the set of categories seen at training~\cite{9577832}. This emerging trend overcomes the limitations affecting long-standing class-specific counting methods, which require individually trained networks for each object type – e.g., people~\cite{8099912,8578662,8954153,10.1007/978-3-030-01228-1_45,8659316,DIBENEDETTO2022117125}, vehicles~\cite{8237658,8969620,CIAMPI2022117929}, or cells~\cite{8265200,CIAMPI2022102500,Ciampi_2022,10619063}. In contrast, CAC draws inspiration from humans’ instinctive ability to discern what merits counting when confronted with unfamiliar objects, and it enables users to specify arbitrary categories of interest at inference time without needing model retraining or new annotated data.

Target object classes in CAC can be specified by users with visual exemplars, \ie, bounding boxes surrounding object samples in the image~\cite{Wang_Xiao_Cao_Lu_2024,DBLP:journals/corr/abs-2405-11770,Liu_2022_BMVC,Dukic_2023_ICCV,10031021}, or with text prompts, \ie, textual object descriptions~\cite{10204688,AminiNaieni23,10.1145/3581783.3611789,Kang_Moon_Kim_Heo_2024,10483595,Pelhan_2024_CVPR,Lin_Chan_2024}. Techniques leveraging visual exemplars currently outperform prompt-based methods, as exemplars offer more detailed information than text, providing inherent information about the visual appearance of the objects. Nevertheless, prompt-based methods enhance flexibility and improve the capabilities and generality of counting models requiring reduced human intervention -- humans do not have to specify bounding boxes as input. These methods leverage the models’ natural language understanding and work hand-in-hand with popular vision-language foundation models such as CLIP~\cite{DBLP:conf/icml/RadfordKHRGASAM21}, thus representing highly appealing solutions~\cite{AminiNaieni23,Pelhan_2024_CVPR}.

However, in this paper, we figured out that current benchmarks exploited for assessing prompt-based CAC approaches are affected by severe limitations hampering their proper evaluation and the development of new, more performing solutions. As shown in~\cref{fig:double-inference}, we empirically verified that some state-of-the-art prompt-based CAC approaches are not always able to truly understand \textit{which} object class has to be counted from the textual description. In contrast, they tend to count object instances belonging to the predominant class despite the prompt meaning. This is a critical failure in real-world scenarios if we imagine that such systems are usually deployed to count specific objects -- \eg, imagine a surveillance scenario in which the officer wants to count the number of pedestrians transiting on the street, but at that moment, only cars are present. We argue that this lack relies on two key factors: (i) the shortcomings affecting the main CAC datasets and (ii) the limitations affecting current counting performance evaluators. Specifically, most of the samples making up the CAC datasets contain objects belonging to a single class, making it hard to evaluate the model’s robustness to discriminate different object classes within an image. Furthermore, the metrics exploited for measuring the performance of CAC models are inherited from class-specific counting and do not consider some key factors of prompt-based CAC, focusing only on the counting error and disregarding assessing the model’s trustworthiness to understand the object category described by the textual prompt.

In this work, we propose \benchmarkname{} (\benchmarkacron{}), a novel benchmark designed to quantitatively evaluate the robustness and trustworthiness of prompt-based CAC approaches, tackling the two above-mentioned limitations. The benchmark includes two tailored tests coupled with ad-hoc evaluation metrics: (i) a \textit{negative-label} test, which probes single-class images by using prompts that refer to absent classes, and (ii) a \textit{mosaic} test, which evaluates artificially created mosaicked images built by stitching together pairs of single-class images, where one object category serves as a distractor to the category described by the textual prompt. To validate our new benchmark, we assess several recent SOTA prompt-based CAC techniques, and we show that some of them show a notable weakness in understanding the object class textual descriptions despite achieving top performance on standard class-specific metrics. This highlights the necessity for more refined training procedures or reconsideration of their architectural designs.

To summarize, we propose the following contributions:
\begin{itemize}
    \item We empirically figured out that the evaluation of state-of-the-art prompt-based class-agnostic counting methods is limited by current datasets and metrics, hindering the development of more effective solutions.
    \item We propose \benchmarkacron{}, a novel benchmark that introduces a set of ad-hoc evaluation protocols and associated metrics to evaluate the robustness and trustworthiness of existing CAC models to count objects belonging to classes described with natural language.
    \item We exploit \benchmarkacron{} to show that many recent state-of-the-art methods exhibit a remarkable deficiency in understanding objects to be counted, although achieving top results in standard class-specific counting metrics.
\end{itemize}

%%%%%%%%%%%%%%%%%%%%%%%%%%%% RELATED WORKS %%%%%%%%%%%%%%%%%%%%%%%%%%%%%%%%%%%%%%%%%%%
\section{Related Work}
\label{sec:related_works}

\subsection{Class-specific Object Counting}
Object counting is one of the core tasks in computer vision due to its widespread applicability to many real-world applications. Thus, many methods have been proposed to count specific object categories, such as people~\cite{8099912,8578662,8954153,10.1007/978-3-030-01228-1_45,8659316}, cars~\cite{8237658,8969620,CIAMPI2022117929}, insects~\cite{BERECIARTUAPEREZ2022106933,CIAMPI2023102384}, or cells~\cite{8265200,CIAMPI2022102500}. Approaches that regress and sum density maps have emerged to be more accurate in crowded scenarios~\cite{8099912,8578662,8954153,10.1007/978-3-030-01228-1_45,8659316,BERECIARTUAPEREZ2022106933} than techniques relying on prior detection of object instances~\cite{8969620,CIAMPI2022117929}. Either way, the main drawback of these methods lies in requiring individually trained networks and, consequently, labeled datasets for each object type.

\subsection{Prompt-based Class-agnostic Counting}
Recently, class-agnostic counting (CAC) has generalized object counting to open-world
settings, where users can specify arbitrary object classes never seen during model training by exploiting (i) visual exemplars, \ie, bounding boxes around the objects of interest within the same input image~\cite{Wang_Xiao_Cao_Lu_2024,DBLP:journals/corr/abs-2405-11770,Liu_2022_BMVC,Dukic_2023_ICCV,10031021}, or (ii) textual prompts, \ie, natural-language descriptions of the object class~\cite{10204688,AminiNaieni23,10.1145/3581783.3611789,Kang_Moon_Kim_Heo_2024,10483595,Pelhan_2024_CVPR,Lin_Chan_2024}. The first setting provides the best performance: methods such as CACViT~\cite{Wang_Xiao_Cao_Lu_2024}, SSD~\cite{DBLP:journals/corr/abs-2405-11770}, CounTR~\cite{Liu_2022_BMVC}, LOCA~\cite{Dukic_2023_ICCV}, and SAFECount~\cite{10031021} reach state-of-the-art in all CAC benchmarks. In contrast, prompt-based CAC approaches are
particularly appealing since they enhance overall flexibility even if they achieve lower performance. 

Most of the SOTA prompt-based CAC approaches rely on vision-language foundation models that map text-image pairs to a joint embedding space. The produced features are then passed to a decoder in charge of producing density maps. 
The authors in~\cite{10204688} trained a conditional variational autoencoder (VAE) based on the popular CLIP model~\cite{DBLP:conf/icml/RadfordKHRGASAM21} to generate visual exemplar prototypes conditioned with their semantic embedding encoded in the provided category name. %; prototypes are then used with existing visual exemplar-based CAC techniques. 
Concurrently,~\cite{AminiNaieni23} proposed CounTX that instead is trained end-to-end and produces object counts directly, skipping the intermediate visual exemplar proposal step. Being based on CLIP to measure the similarity between image patches and class descriptions, it also accepts a more detailed specification of the target object to count (rather than simply using a class name). %More in detail, it is built on top of a CLIP model, where a feature interaction module is used to mine the similarities between image patches and class descriptions.
Similarly, CLIP-Count~\cite{10.1145/3581783.3611789} relies on CLIP to align the text embedding with dense visual features. However, the authors also designed a hierarchical patch-text interaction module to propagate semantic information across different resolution levels of visual features. Another similar CLIP-based model trained end-to-end is VLCounter~\cite{Kang_Moon_Kim_Heo_2024}. Here, the authors incorporated three modules to efficiently finetune CLIP for the counting task and to exploit intermediate features across different encoding layers of CLIP in the decoding stage. 
Very recently, DAVE~\cite{Pelhan_2024_CVPR} proposed a two-stage detect-and-verify paradigm. In the first stage, they estimate a high-recall set of candidate detections, which are then analyzed and filtered in the second verification step, relying on unsupervised clustering and CLIP embedding. Differently from the above-described architectures, TFPOC~\cite{10483595} employed a detection-based technique, leveraging the popular SAM~\cite{kirillov2023segment} for instance segmentation. The authors proposed a two-stage approach. In the first stage, they exploited CLIP-Surgery~\cite{DBLP:journals/corr/abs-2304-05653}, an enhanced version of CLIP, to produce visual exemplar prototypes by computing the similarity between images and text representations. In the second step, they computed a similarity map between image features and the masks corresponding to the reference objects produced by SAM prompted with the bounding boxes from the first stage. %However, in this work, we empirically demonstrate that these prompt-based CAC models are not always able to truly understand object categories to be counted from textual descriptions, often defaulting to the most prevalent object classes instead. 
%This stems also from the fact that prompt-based approaches emerged after the ones based on visual exemplars and they inherited their setting: the provided textual descriptions usually refer to object classes that are grounded in the image, as visual exemplars localize only objects present in the image itself. We relax this requirement, considering a real-world scenario where the user can provide textual descriptions of  

All of these prompt-based methods derive from the literature on exemplar-based CAC, where elements to be counted are chosen by selecting some exemplars from the given image. As a result, prompt-based CAC methods also assume the object is present in the image. Our goal is to test a scenario where this assumption is relaxed, allowing for the possibility that (i) the object may not be present in the image, (ii) the image may contain more than one object class, or even (iii) the user provides a misleading or ambiguous query. We empirically demonstrate that prompt-based CAC models are activated by non-present categories and, therefore, are unable to truly understand object categories to be counted from textual descriptions.

%A Fixed-Point Approach to Unified Prompt-Based Counting~\cite{Lin_Chan_2024} --> NO CODE, non lo considered

%Some interesting papers:
%\begin{itemize}
%    \item LAMM: Language-Assisted Multi-Modal Instruction-Tuning Dataset, Framework, and Benchmark - NIPS 2023 \cite{NEURIPS2023_548a41b9} -- useful for eventually take results obtained with multimodal LLMs
%\end{itemize}

\subsection{Dataset and Metrics}
The gold standard dataset for CAC is FSC-147~\cite{9577832}. It contains 6,135 images across 147 object categories, with 89 categories used for training, 29 for validation, and 29 for testing, with no class overlap between these subsets. Annotations consist of dots over the approximate centroids of each object instance, as usual for the counting task -- ground truth density maps are created by superimposing Gaussian kernels centered on these dots. Furthermore, the authors provided three bounding boxes per image that localize the exemplars and the natural language name of the object category to which they belong – the category is unique for each image, and it has been chosen arbitrarily if multiple categories were present. However, cases in which multiple object categories are present in the same image are just a few, and this represents an inherent limitation of FSC-147, making it hard to evaluate the model’s robustness to discriminate different object classes within an image. To address this shortcoming, two other datasets
have been proposed: OmniCount-191~\cite{DBLP:journals/corr/abs-2403-05435} and MCAC~\cite{10.1007/978-3-031-73247-8_18}. These datasets include multiple object categories within the same images. However, the first one is not publicly available, while MCAC lacks suitable annotations for prompt-based approaches.

Standard counting metrics in the literature are the mean absolute error (MAE) and the root mean squared error (RMSE), defined as $\text{MAE} = \frac{1}{N} \sum_{n=1}^{N} \left| \tilde{c}_n - c_n \right|$ and $\text{RMSE} = \sqrt{\frac{1}{N} \sum_{n=1}^{N} ( \tilde{c}_n - c_n )^2}$, where $N$ is the number of test images, and $\tilde{c}_n$ and $c_n$ are the ground truth and predicted counts, respectively. Another performance evaluator, used less frequently, is represented by the mean absolute percentage error (MAPE), which is essentially a normalized MAE and is defined as $\text{MAPE} = \frac{1}{N} \sum_{n=1}^{N} \frac{|\tilde{c}_n - c_n|}{\tilde{c}_n}$. However, these evaluators were inherited from the class-specific counting task and exhibit severe limitations particularly evident in prompt-based CAC. Specifically, they do not take into account \textit{which} object category has to be counted nor the model's ability to understand the provided textual prompt. 

%In this work, we introduce a novel counting benchmark specific for prompt-based CAC that overcomes the limitations due to current datasets and performance evaluators.

\begin{figure*}[t]
    \centering
    \includegraphics[width=1\linewidth,page=1]{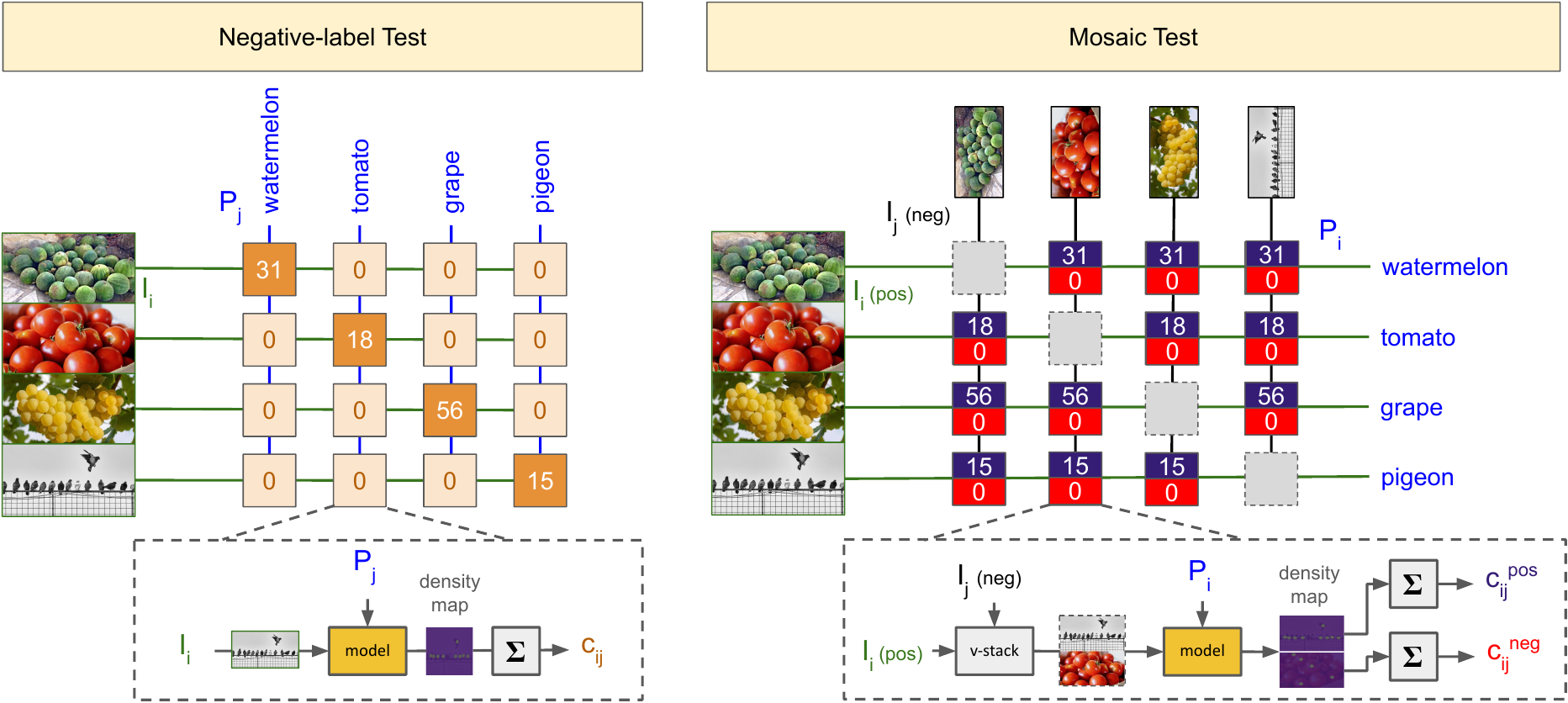}
    \caption{Inference schemas for the \textit{negative-label test} (on the left) and the \textit{mosaic test} (on the right). The numbers reported in the boxes show the ideal model outcomes: for the negative-label test, the diagonal of the shown matrix should be filled with the ground-truth object counts and with zeros elsewhere; for the mosaic test, each mosaic outcome should contain the ground-truth counts on the top -- which is the same for each row of the shown matrix -- and zeros on the bottoms. In the dotted boxes on the bottom, we report a schema of the inference procedures needed for computing each entry of the two matrices.}
    \label{fig:inference}
\end{figure*}

\section{The \benchmarkacron{} Benchmark}
%In this section, we showcase our novel \benchmarkname{} (\benchmarkacron{}) benchmark, which introduces ad-hoc evaluation protocols and metrics underlining the limitations of counting models to correctly interpret the provided textual prompts.

\subsection{Overview}

We assume to have a collection of $N$ tuples $\mathcal{X} = \{(I_1, P_1), (I_2, P_2), \dots, (I_N, P_N)\}$, where $I_i$ is an image and $P_i$ is the class name of the objects within $I_i$, \ie, the description of the object class present in it. Thus, each image contains objects belonging to only a single class.

We define $P_i$ as the \textit{positive} class and all the $\{P_j\}_{j\neq i}$ as the \textit{negative} classes for the given image $I_i$. Without loss of generality, we also assume that each class is represented by only one image in the dataset, even though, in real-world scenarios, datasets typically contain multiple images of the same object class. Finally, we assume to have a prompt-based CAC model $\mathcal{M}(I, P)$, which takes as input a generic image $I$ together with an arbitrary prompt $P$ and outputs an estimated count $c$ of the object instances belonging to the object class described by $P$.
In this setting, \benchmarkacron{} introduces two test suites, which are detailed in the following sections: (i) a \textit{negative-label} test and (ii) a \textit{mosaic} test.

% In light of the above, in this paper, we propose a novel counting benchmark, which we call \benchmarkname{}, which will help in understanding to which extent current models are able to understand the input textual prompt, providing correct counting estimate for the requested class while ignoring the others. 

%The \benchmarkacron{} benchmark is composed of two distinct tests. The first test, also named \textbf{Negative-label test}, aims to understand how the model behaves when a label of an object belonging to a class not present in the query image is provided as the textual prompt. The second test, named \textbf{Mosaic test}, aims to understand how the model behaves when two classes are present in the scene and only one of them is required to be counted by the textual prompt.
%We describe them in detail in the following sections.

\subsection{Negative-label test}
The \textit{negative-label} test evaluates single-class images by prompting the model with textual descriptions referring to absent object classes. 
%Formally, 
%The Negative-label test aims to evaluate single-class images by prompting the model with textual description referring to absent object classes, thus testing 
%The Negative-label test aims to test 
%whether counting models pay attention to categories not present in query images. %This test is prepared as follows. 
% where each image $I_i$ include only objects belonging to a single \textit{positive} class $P_i$, 
%We suppose every image $\mathbf{I}_i$ in the dataset has a single positive class $p^\text{pos}_i$ -- which is the class of the sole objects present in the image -- 
%while the other $k$ classes within the dataset dictionary $\{p_{ij}^\text{neg}\}_{j=1}^k$ are considered negative classes for that image.
Formally, for all the elements $\{I_i, P_j\}_{i,j=1}^N$, we compute $c_{ij} = \mathcal{M}(I_i, P_j)$ to obtain the predicted count for each image when prompted with all the available class descriptions in the dataset. The optimal test outcome can be formalized as follows:
\begin{equation}
    c_{ij} =
    \begin{cases}
        \tilde{c_i}, & \text{if } i = j \\
        0, & \text{if } i \neq j
    \end{cases},
\end{equation}

\noindent where $\tilde{c}_i$ is the ground truth count for the image $I_i$.

The ideal situation is illustrated in the left part of~\cref{fig:inference}. Intuitively, standard class-specific counting metrics consider only the count predictions concerning the positive class -- the orange diagonal in the matrix of~\cref{fig:inference}. Conversely, we account for the count predictions concerning the negative classes -- the remaining non-diagonal elements in the matrix of~\cref{fig:inference}. To quantitatively assess this test, we introduce two ad-hoc
metrics.

\paragraph{Normalized Mean of Negative predictions (NMN).} 
NMN is the absolute counting error computed by prompting the model with the negative classes, normalized by the ground truth of the positive class. % average of the ratios of the mean of predictions obtained with each label of the negative classes as textual prompt and the ground truth count. 
It is computed as follows (more details in the supplementary material):
\begin{equation}
\text{NMN} = \frac{1}{N(N-1)} \sum_{i=1}^{N} \frac{1}{\tilde{c}_i} \sum_{\substack{j=1 \\ j \neq i}}^{N} c_{ij}
\end{equation}
Notice that the reason for normalizing the average of the negative counts by the ground truth of the positive class is given by the assumption that having more items in the image also causes the model to increase the estimate of non-present classes. Therefore, the normalization factor $1 / \tilde{c}_i$ plays the role of relativizing the error -- in other words, we suppose that counting 10 \textit{dogs} in an image showing 1000 \textit{people} is a negligible error.
%This metric is interesting to understand if the model can ignore objects in the image if a negative class label is provided as the textual prompt. 
The lower the NMN value, the better the resulting model.

\paragraph{Positive Class Count Nearness (PCCN).} 
PCCN mixes positive and negative class predictions, providing an overall quantitative assessment of strong failures in the model. Formally, it is defined as follows:
\begin{equation}
\text{PCCN} = \frac{1}{N} \sum_{i=1}^{N} \mathbb{I}(d^{\text{pos}}_i < d^{\text{neg}}_i) \cdot 100\%
\label{eq:wpr}
\end{equation}

\noindent where $d^{\text{neg}}_i = | \frac{1}{N} \sum_{\substack{j=1 \\ j \neq i}}^N c_{ij} - \tilde{c}_i |$ is the absolute difference between the mean of the model's negative classes predictions and the ground truth for the $i$-th image, $d^{\text{pos}}_i = | c_{ii} - \tilde{c_i} |$ is the absolute difference between the model's positive class prediction and the ground truth count for the $i$-th image, and $ \mathbb{I}(\cdot)$ is an indicator function, which equals 1 if the condition inside it is true, and 0 otherwise. Thus, PCCN measures the percentage of data samples for which the model produces a positive class count estimate that is closer to the ground truth compared to the mean of the negative class count estimations. Intuitively, a mean of the negative class estimations that is closer to the ground truth than a positive class estimate may indicate that the model is strongly biased toward counting negative classes over the positive class, which in turn may be due to the model completely losing the semantics of the given prompt. It follows that the higher the PCCN, the better the resulting model.

\subsection{Mosaic test}
The \textit{mosaic} test assesses images containing multiple object categories by providing the model with textual prompts referring only to the positive class, emulating very common real-case scenarios. Thus, the object instances belonging to the negative classes serve as distractors to the positive class. To overcome the lack of suitable annotated CAC datasets having multi-class images, we artificially build a new set of such images. Specifically, starting from existing single-class CAC datasets, we create mosaicked images by stitching together pairs of positive and negative single-class images, similarly to~\cite{AminiNaieni23}.

Formally, we consider all the possible pairs of images $\{I_i, I_j\}_{\substack{i,j=1 \\ j \neq i}}^N$, and the positive class $P_i$ of the image $I_i$. We call $I_i$ and $I_j$ the \textit{positive} and the \textit{negative} images, respectively. We create new mosaicked images $I_{ij}^\text{mosaic} = \text{vstack}(I_i, I_j)$ by vertically stitching together the positive image, placed at the top, with the negative one, placed at the bottom. Then, the CAC model $\mathcal{M}$ is tasked to predict the following quantities concerning the positive class $P_i$:
\begin{equation}
    c_{ij}^\text{pos}, c_{ij}^\text{neg} = \mathcal{M}(I_{ij}^\text{mosaic}, P_i),
\end{equation}
where $c_{ij}^\text{pos}$ and $c_{ij}^\text{neg}$ are the positive and negative class counts relative to the top and the bottom parts of the mosaic, respectively. This can be easily obtained from most density-based models, as it is sufficient to cut the output density map in half along the y-axis and integrate them separately to obtain the respective positive and negative counts.
It is worth noting that negative images influence not only the negative class count $c_{ij}^\text{neg}$, but also the positive class count $c_{ij}^\text{pos}$. Indeed, the goal of this test is to evaluate the robustness of the model to count the positive class described by the prompt in the presence of distractors, \ie, objects from different categories within the same image.
%In this test, note that not only does the negative count $c_{ij}^\text{neg}$ depend on the negative images, but the positive count $c_{ij}^\text{pos}$ is also influenced by them. Indeed, the presence of different negative images in the lower part of the mosaic can distract the model, affecting its accuracy in counting positive instances.
Similarly to the negative-label test, we ideally want $c_{ij}^\text{pos}$ to converge to the ground truth count, while $c_{ij}^\text{neg}$ should ideally approach zero. This means the model (i) should not be distracted by negative examples when attending to positive instances, and (ii) should refrain from counting any negative instances in the mosaic. We summarize this inference procedure in the right part of~\cref{fig:inference}.

Since each mosaic includes both positive and negative instances, this test bears similarities to object detection evaluation, where both correct and incorrect detections can occur, and performance is typically assessed using \textit{precision}, which reflects the proportion of correct predictions among all model outputs, and \textit{recall}, which represents the percentage of correct instances retrieved from the total ground-truth.
%therefore a group of metrics commonly exploited in these tasks are replicated. 
%These metrics are similar to Precision, Recall and Balanced F-score, but 
However, differently from object detection, in our scenario, we do not have clear indications about the correctness of each proposed instance, given that we do not know -- and we are not requested to know -- precise instance localization within the image, as we are only interested in the final aggregated count for the provided object class. % For this reason, it is difficult to derive true/false positives/negatives to assemble precision and recall metrics. 
Such information could be partially recovered from the two output counts $c_{ij}^\text{pos}$ and $c_{ij}^\text{neg}$ by making some reasonable assumptions that we present in the following, where we introduce precision-recall metrics shaped for prompt-based CAC.

\begin{figure}[t]
    \centering
    \includegraphics[width=.8\linewidth,page=3]{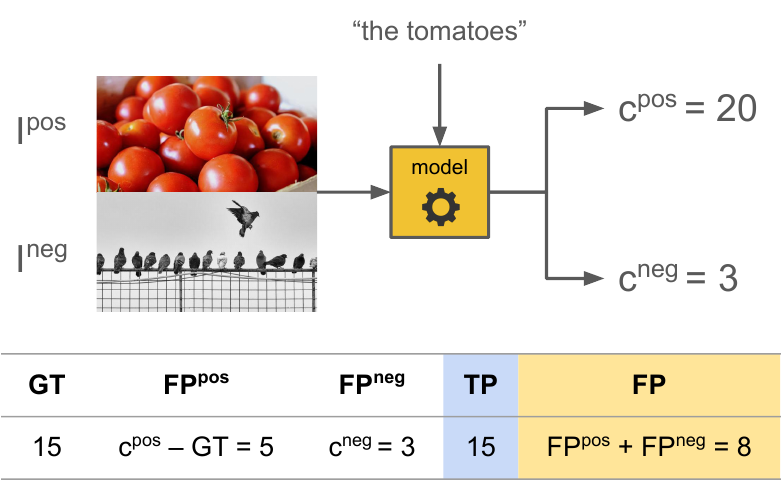}
    \caption{Example of the derivation of TPs and FPs in the mosaic scenario. In the shown case where the model predicts $c^\text{pos} = 3$ and $c^\text{neg} = 20$, the number of estimated true positives is bounded to the ground-truth value (15). The remaining 5 counted elements are considered false positives from the positive image ($\text{FP}^\text{pos} = 5$), which are then merged with the false positives from the negative image ($\text{FP}^\text{neg} = c^\text{neg} = 3$), to obtain a total of 8 FPs.}
    \label{fig:example-mosaic}
\end{figure}

\paragraph{Counting Precision (CntP) and Counting Recall (CntR).}
Classical precision and recall metrics used in detection scenarios are defined through true positive count (TP), false positive count (FP), and false negative count (FN).
In the counting scenario, we do not have an indication of whether a single instance is correct or not, as we only have the estimated global count for the provided class of objects. We only know that correct objects are only present in the positive image $I_i$ and that the negative image $I_j$ only contains incorrect examples for the prompt $P_i$. %\footnote{In the remainder of this paragraph, we will focus on a single mosaic image. Therefore, we will simplify the notation by excluding $i$ and $j$ indexes.} 
To adapt precision and recall metrics to the counting case, we make a simple assumption. Specifically, for the counting contributions $c_{ij}^\text{pos}$ from the positive image $I_i$, we never have FN instances. We may only have FPs in the case where the predicted count $c_{ij}^\text{pos} > \tilde{c}_i$, where $\tilde{c}_i$ is the ground truth count for the positive instances in $I_i$. In such a case, we are assuming that the instances exceeding the correct count are false predictions. Also note that all the contributions $c_{ij}^\text{neg}$ coming from the negative image $I_j$ are surely FPs, as in such cases, the model is erroneously counting the incorrect instances in the lower part of the mosaic. \Cref{fig:example-mosaic} shows an example of the computation of TPs and FPs for adapting the precision and recall detection metrics to the counting scenario.

Under the above considerations and assumptions, we obtain that \textit{counting precision} (CntP) and \textit{counting recall} (CntR) can be defined as follows:
\begin{align}
    \label{eq:counting_precision}
    \text{CntP} &= \frac{1}{N(N-1)}\sum_{\substack{i,j=1 \\ j \neq i}}^N \frac{\min(c_{ij}^\text{pos}, \tilde{c}_i)}{c_{ij}^\text{pos} + c_{ij}^\text{neg}} \\
    \text{CntR} &= \frac{1}{N(N-1)}\sum_{\substack{i,j=1 \\ j \neq i}}^N\frac{\min(c_{ij}^\text{pos}, \tilde{c}_i)}{\tilde{c}_i}
\end{align}

We refer the reader to the supplementary material for the derivation of such expressions from the idea in \cref{fig:example-mosaic}.

The \textit{counting recall} metric is useful to understand if the model can correctly estimate the number of objects belonging to the positive class. Interestingly, in the range $c_{ij}^\text{pos} < \tilde{c}_i$, it correlates negatively with the MAPE error metric -- clamping to 1 when the predicted count is greater than the ground truth. Therefore, CntR carries information partially correlated with class-specific counting metrics, except that the model is also exposed to negative examples that may alter the number of predicted instances.

Differently, the \textit{counting precision} metric is useful to understand if the model is erroneously including in the count estimation also objects belonging to classes not specified by the textual prompt and present in the negative image $I_j$. Specifically, it is sensitive to the FPs found both in the positive image $I_i$ -- the counting excess with respect to the ground-truth -- and in the negative image $I_j$ -- where every instance is considered a FP.

\paragraph{Counting Balanced F1-score (CntF1).}

A common way to aggregate precision and recall to obtain a single indicator is the F1-score, which is defined as the harmonic mean of the two metrics. We therefore derive a \textit{counting F1-score} (CntF1) indicator, as follows:

\begin{equation}
\text{CntF1} = 2\cdot\frac{\text{CntP} \cdot \text{CntR}}{\text{CntP} + \text{CntR}}
\end{equation}

This metric is useful for comparing the overall performance of the CAC models, taking into consideration objects belonging both to desired and undesired classes.

\begin{table*}[t]
    \centering
    \caption{Results on the \textbf{test set} of FSC147 on both \benchmarkacron{} metrics (negative and mosaics tests) and on classic metrics.}%
    \label{tab:results-FSC147-test-set}%
    \newcolumntype{L}{>{\arraybackslash}m{.25\linewidth}}%
    \newcolumntype{C}{>{\centering\arraybackslash}X}
    %\tiny%
    \setlength{\tabcolsep}{1pt}
    \begin{tabularx}{0.95\linewidth}{L|CC|CCC|CC}
        \toprule
        & \multicolumn{2}{c|}{Negative Test} & \multicolumn{3}{c|}{Mosaic Test} & \multicolumn{2}{c}{Classic} \\
        \cmidrule(lr){2-3} \cmidrule(lr){4-6} \cmidrule(lr){7-8}
        \centering Method & NMN $\downarrow$ & PCCN $\uparrow$ & CntP $\uparrow$ & CntR $\uparrow$ & CntF1 $\uparrow$ & MAE $\downarrow$ & RMSE $\downarrow$ \\
        \midrule
        \midrule
        CounTX~\cite{AminiNaieni23} {\footnotesize (BMVC '23)} & 0.95	& 64.51 & 0.686 & 0.712 & 0.630 & 15.92 & 106.89 \\
        CLIP-Count~\cite{10.1145/3581783.3611789} {\footnotesize (ACM MM '23)} & 1.27 & 38.13 & 0.495 & 0.761 & 0.554 & 17.59 & 109.97 \\
        VLCounter~\cite{Kang_Moon_Kim_Heo_2024} {\footnotesize (AAAI '24)} & 1.15 & 53.36 & 0.517 & 0.781 & 0.577 & 17.02 & 106.93 \\
        TFPOC~\cite{10483595} {\footnotesize (WACV '24)} & 0.75 & 66.04 & 0.687 & \textbf{0.848} & 0.696 & 24.79 & 138.11 \\
        DAVE~\cite{Pelhan_2024_CVPR} {\footnotesize (CVPR '24)} & \textbf{0.08} & \textbf{97.62} & \textbf{0.843} & 0.799 & \textbf{0.790} & \textbf{15.23} & \textbf{103.53} \\
        \bottomrule
    \end{tabularx}
\end{table*}

\begin{table*}[t]
    \centering
    \caption{Results on the \textbf{validation set} of FSC147 on both \benchmarkacron{} metrics (negative and mosaics tests) and on classic metrics.}
    \label{tab:results-FSC147-val-set}%
    \newcolumntype{L}{>{\arraybackslash}m{.25\linewidth}}%
    \newcolumntype{C}{>{\centering\arraybackslash}X}
    %\tiny%
    \setlength{\tabcolsep}{1pt}
    \begin{tabularx}{0.95\linewidth}{L|CC|CCC|CC}
        \toprule
        & \multicolumn{2}{c|}{Negative Test} & \multicolumn{3}{c|}{Mosaic Test} & \multicolumn{2}{c}{Classic} \\
        \cmidrule(lr){2-3} \cmidrule(lr){4-6} \cmidrule(lr){7-8}
        \centering Method & NMN $\downarrow$ & PCCN $\uparrow$ & CntP $\uparrow$ & CntR $\uparrow$ & CntF1 $\uparrow$ & MAE $\downarrow$ & RMSE $\downarrow$ \\
        \midrule
        \midrule
        CounTX~\cite{AminiNaieni23} {\footnotesize (BMVC '23)} & 0.87	& 69.79 & 0.664 & 0.658 & 0.579 & 17.27 & 66.37 \\
        CLIP-Count~\cite{10.1145/3581783.3611789} {\footnotesize (ACM MM '23)} & 1.24 & 48.11 & 0.488 & 0.702 & 0.522 & 18.90 & 66.75 \\
        VLCounter~\cite{Kang_Moon_Kim_Heo_2024} {\footnotesize (AAAI '24)} & 1.07 & 62.70 & 0.520 & 0.741 & 0.561 & 18.09 & 65.59 \\
        TFPOC~\cite{10483595} {\footnotesize (WACV '24)} & 0.67 & 62.62 & 0.723 & \textbf{0.757} & 0.656 & 32.84 & 110.60 \\
        DAVE~\cite{Pelhan_2024_CVPR} {\footnotesize (CVPR '24)} & \textbf{0.13} & \textbf{95.90} & \textbf{0.819} & 0.751 & \textbf{0.757} & \textbf{16.58} & \textbf{54.73} \\
        \bottomrule
    \end{tabularx}
\end{table*}

\section{Experimental Evaluation}

\subsection{Dataset and Methods}

\paragraph{Dataset.}
Our benchmark relies on FSC-147~\cite{9577832}, a widely used dataset for CAC containing 6,135 images with objects belonging to 147 classes. Labels include dots over the object centroids, bounding boxes for three object exemplars, and object category textual names (see also \cref{sec:related_works}). Although most images contain objects belonging to a single category, we filter out the few multi-class images by following the approach in~\cite{Pelhan_2024_CVPR}. This prevents interferences with our proposed tests, ensuring that no false positives arise from other object classes present in the same image.

\paragraph{Probed Methods.}
We place under the spotlight six different state-of-the-art prompt-based CAC methods: CounTX~\cite{AminiNaieni23}, CLIP-Count~\cite{10.1145/3581783.3611789}, TFPOC~\cite{10483595}, VLCounter~\cite{Kang_Moon_Kim_Heo_2024}, and DAVE~\cite{Pelhan_2024_CVPR}. 
%and CountGD [].
Specifically, CounTX, CLIP-Count, and VLCounter directly estimate density-maps end-to-end by fine-tuning CLIP and conditioning the density-map generation on the CLIP embedding of the desired object class. Instead, TFPOC and DAVE employ two-stage approaches that detect all the objects in the image and then filter them based on the input textual prompt.

\paragraph{Implementation Details.}
We employed the original code and pre-trained models provided by the authors, maintaining their image pre-processing pipelines, hyperparameters, and
prompting schema. We only needed to adjust the DAVE inference procedure to behave correctly on our benchmark. In the supplementary material, we provide further details on these changes along with the \benchmarkacron{} performance of the original DAVE implementation.

\begin{figure}[t]
    \centering
    \includegraphics[width=\linewidth]{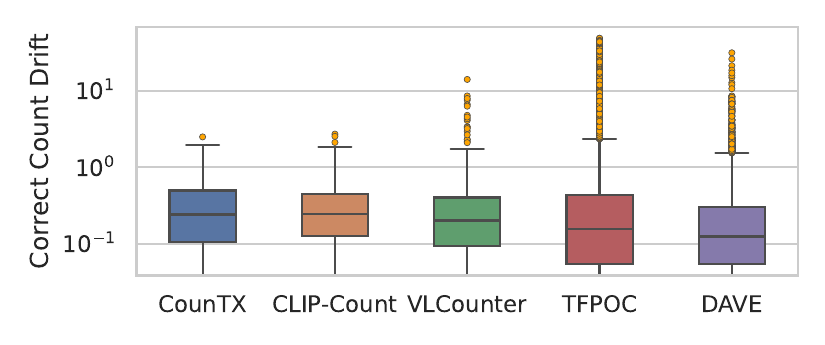}
    \caption{Boxplot showing the distribution of the correct count drifts of the different models. Despite the lower mean value, TFPOC and DAVE show a consistent number of outliers, revealing that they may catastrophically fail in some specific conditions.}
    \label{fig:correct-count-drift}
\end{figure}

\begin{figure*}[t]
    \centering
    \includegraphics[width=0.9\linewidth,trim={1.1cm 1.1cm 0 0},clip]{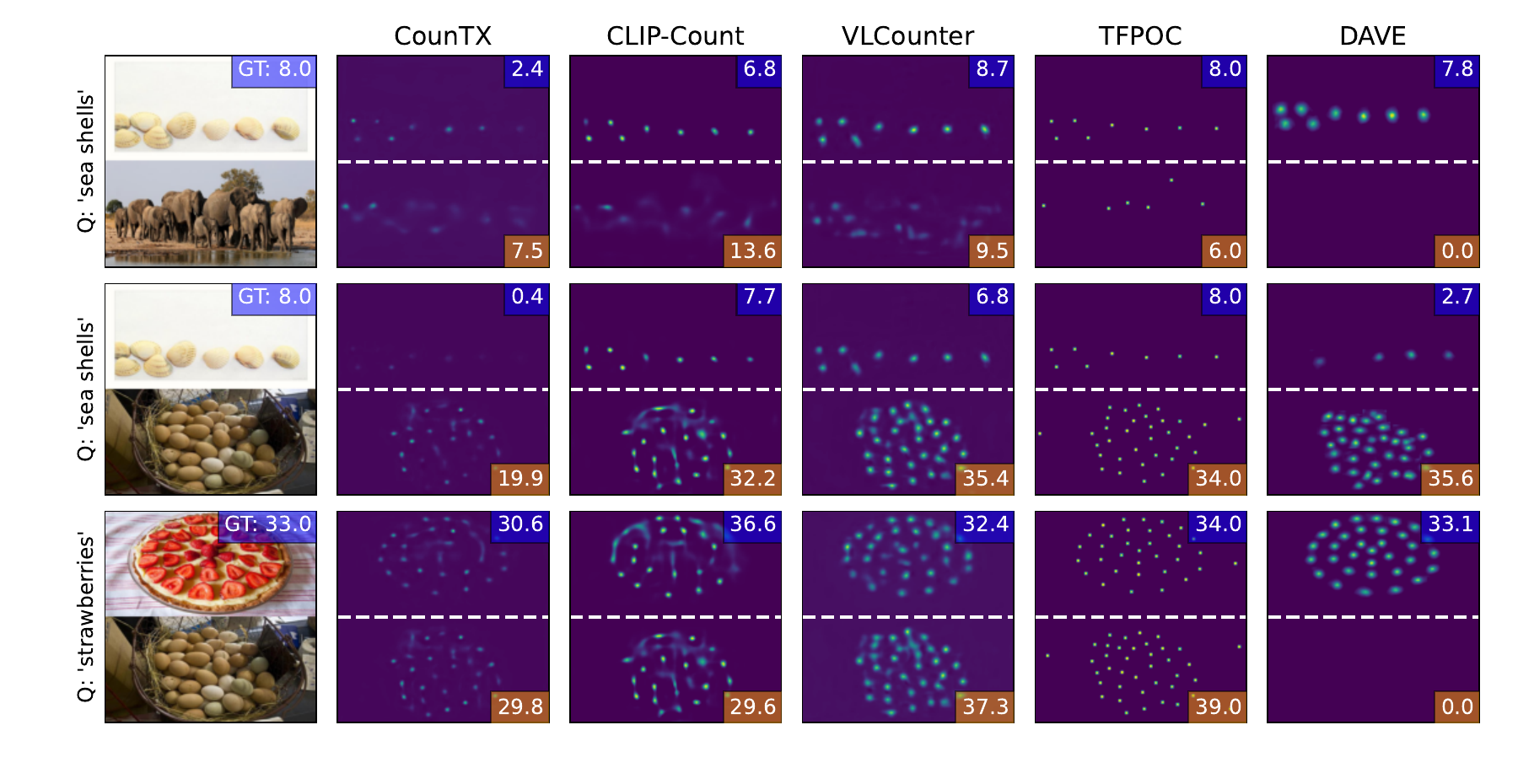}
    \caption{This figure shows, for each model, the output density maps for three different \textit{(mosaic, input prompt)} pairs. The count reported in the blue box is $c^\text{pos}_{ij}$, while the count reported in the dark orange box corresponds to $c^\text{neg}_{ij}$. We can notice how the models often misidentify instances from the negative image in the mosaic, though most accurately estimate the positive instances in the upper part.}
    \label{fig:qualitative-results-mosaic}
\end{figure*}

\subsection{Results}

\paragraph{Quantitative Results on Negative and Mosaic Tests.}
We report the results for both the test and validation splits in~\cref{tab:results-FSC147-test-set} and~\cref{tab:results-FSC147-val-set}, respectively.
As we can notice, although the methods achieve remarkable performance on standard counting error measures (MAE and RMSE), the outcome is quite heterogeneous on the \benchmarkacron{} metrics. Notably, the negative test on one-stage methods like CounTX, CLIP-Count, and VLCounter shows that the average negative count is comparable to the ground-truth count of the correct class ($\text{NMN} \approx 1$), with CLIP-Count and VLCount achieving $\text{NMN} > 1$. This trend is further confirmed by the PCCN metric, where CLIP-Count has the correct estimate nearer to the ground-truth only 38\% of the times on the test set. Instead, the two-stage detectors TFPOC and DAVE achieve the best performance on the negative test. The same trends are mostly confirmed on the mosaic test. Specifically, while the CntR metric only increments by around 12\% between the worst-performing method (CounTX) and the best-performing one (DAVE), the CntP shows a notable increase, from 0.517 of VLCounter to 0.843 of DAVE, which is an improvement of around 63\%. This shows how the methods, despite being mostly able to count the correct class, have a highly heterogeneous behavior with respect to the negative images within the mosaics. It is also worth noticing that more recent methods do not always improve on the \benchmarkacron{} metrics, meaning that this aspect is still largely underestimated. It is interesting to observe how TFPOC, a training-free model obtaining less competitive results on classic counting metrics, can instead defeat older learning-based CAC models on the proposed metrics.

\paragraph{Correct Count Drift.}
From the mosaic test, it is also interesting to understand how the various models drift the estimate about the positive class $c^\text{pos}_{ij}$ when changing the negative image $I_j$. In other words, we would like to estimate the \textit{confusion} of the model when it is requested to count a given class, but objects of another class are present in the image. %there are objects of another class in the image. 
In~\cref{fig:correct-count-drift}, we report, for each model, the distribution of the quantity $\Bigl\{\frac{|c^\text{pos}_{ij} - c_{ii}|}{c_{ii}}\Bigl\}_{i,j=1}^N$, which encodes the normalized absolute error between the estimate of the model when only presented with the correct class as estimated in the negative test $(c_{ii})$ and the estimate of the model when presented with all the mosaics constructed with positive image $I_i$ on the top ($c^\text{pos}_{ij}$).
As we can notice, although DAVE achieves the lowest drift, it also presents many possibly problematic outliers. This may be due to the verification module, which incorrectly assigns the positive label to the cluster obtained from the objects from the negative image. Despite having a higher mean drift, the other methods -- especially CounTX and CLIP-Count -- show fewer outliers. This signifies that, although these methods wrongly count the negative classes, the count relative to the positive part of the mosaic is less dependent on the number of negative instances from the bottom image of the mosaic. This insight further raises the attention to effective two-stage methods that, despite being very effective in discriminating the different classes, may catastrophically fail in some specific scenarios.

\paragraph{Qualitative Results.}
In~\cref{fig:qualitative-results-mosaic}, we show some qualitative results from the various methods examined through the lenses of the mosaic test. The lack of a proper understanding of the correct class to count is largely deducible from the reported density maps. While many methods correctly estimate the positive class on the top of the mosaics, most of the models give a non-zero count prediction of the negative instances from the negative images.
We can notice how DAVE, in the first and last rows, correctly ignores the negative instances, exactly estimating zero objects. However, its second-stage verification approach may occasionally fail in a catastrophic manner, assigning the label to all the instances of the negative class, as shown in the second reported example. Although the mosaic test already gives many insights into the failure modes of the probed SOTA models, in the supplementary material, we also report some qualitative results from the negative test.

\section{Conclusions}

In this paper, we introduced \benchmarkname{} (\benchmarkacron{}), a novel benchmark composed of two targeted test suites specifically designed to address the limitations of current evaluation systems for prompt-based CAC. %\benchmarkacron{} consists of two distinct tests: a negative test, which evaluates the model's response to prompts referring to absent classes, and a mosaic test, which assesses the ability to count objects from a specified class within multi-class images. To complement these tests, we proposed new evaluation metrics that provide a more comprehensive assessment of prompt understanding, beyond traditional counting errors.

Our evaluation of several recent state-of-the-art CAC methods revealed significant performance gaps, particularly in the ability of the models to handle negative and multi-class scenarios. While two-stage models like DAVE generally outperformed one-stage models on the negative test, they still exhibited notable weaknesses, particularly in the mosaic test, where they sometimes struggled with false positives from the negative images. %One-stage models like CounTX and CLIP-Count showed fewer outliers in terms of positive class drift but continued to incorrectly count negative classes.

These findings suggest that SOTA models, despite their success on standard metrics, require more careful designs and better training procedures to improve their understanding of textual prompts. We hope \benchmarkacron{} provides a foundation for a more comprehensive evaluation and opens the door for developing more robust methods in the near future.

\section*{Acknowledgments}
This work was partially funded by: Spoke 8, Tuscany Health Ecosystem (THE) Project (CUP B83C22003930001), funded by the National Recovery and Resilience Plan (NRRP), within the NextGeneration Europe (NGEU) Program; SUN -- Social and hUman ceNtered XR (EC, Horizon Europe No. 101092612); PNRR-M4C2 (PE00000013) "FAIR-Future Artificial Intelligence Research" - Spoke 1 "Human-centered AI", funded under the NextGeneration EU program; Italian Ministry of Education and Research (MUR) in the framework of the \texttt{FoReLab} projects (Departments of Excellence).

%%%%%%%%% REFERENCES
{\small
\bibliographystyle{ieee_fullname}
\bibliography{main}

\begin{thebibliography}{10}\itemsep=-1pt

\bibitem{8969620}
Giuseppe Amato, Luca Ciampi, Fabrizio Falchi, and Claudio Gennaro.
\newblock Counting vehicles with deep learning in onboard uav imagery.
\newblock In {\em 2019 IEEE Symposium on Computers and Communications (ISCC)}, pages 1--6, 2019.

\bibitem{AminiNaieni23}
N. Amini-Naieni, K. Amini-Naieni, T. Han, and A. Zisserman.
\newblock Open-world text-specified object counting.
\newblock In {\em BMVC}, 2023.

\bibitem{BERECIARTUAPEREZ2022106933}
Arantza Bereciartua-Pérez, Laura Gómez, Artzai Picón, Ramón Navarra-Mestre, Christian Klukas, and Till Eggers.
\newblock Insect counting through deep learning-based density maps estimation.
\newblock {\em Computers and Electronics in Agriculture}, 197:106933, 2022.

\bibitem{10.1007/978-3-030-01228-1_45}
Xinkun Cao, Zhipeng Wang, Yanyun Zhao, and Fei Su.
\newblock Scale aggregation network for accurate and efficient crowd counting.
\newblock In Vittorio Ferrari, Martial Hebert, Cristian Sminchisescu, and Yair Weiss, editors, {\em ECCV}, pages 757--773, Cham, 2018. Springer International Publishing.

\bibitem{Ciampi_2022}
Luca Ciampi, Fabio Carrara, Giuseppe Amato, and Claudio Gennaro.
\newblock Counting or localizing? evaluating cell counting and detection in microscopy images.
\newblock In {\em Proceedings of the 17th International Joint Conference on Computer Vision, Imaging and Computer Graphics Theory and Applications}, page 887–897. SCITEPRESS - Science and Technology Publications, 2022.

\bibitem{CIAMPI2022102500}
Luca Ciampi, Fabio Carrara, Valentino Totaro, Raffaele Mazziotti, Leonardo Lupori, Carlos Santiago, Giuseppe Amato, Tommaso Pizzorusso, and Claudio Gennaro.
\newblock Learning to count biological structures with raters’ uncertainty.
\newblock {\em Medical Image Analysis}, 80:102500, 2022.

\bibitem{CIAMPI2022117929}
Luca Ciampi, Claudio Gennaro, Fabio Carrara, Fabrizio Falchi, Claudio Vairo, and Giuseppe Amato.
\newblock Multi-camera vehicle counting using edge-ai.
\newblock {\em Expert Systems with Applications}, 207:117929, 2022.

\bibitem{CIAMPI2023102384}
Luca Ciampi, Valeria Zeni, Luca Incrocci, Angelo Canale, Giovanni Benelli, Fabrizio Falchi, Giuseppe Amato, and Stefano Chessa.
\newblock A deep learning-based pipeline for whitefly pest abundance estimation on chromotropic sticky traps.
\newblock {\em Ecological Informatics}, 78:102384, 2023.

\bibitem{8265200}
Joseph~Paul Cohen, Geneviève Boucher, Craig~A. Glastonbury, Henry~Z. Lo, and Yoshua Bengio.
\newblock Count-ception: Counting by fully convolutional redundant counting.
\newblock In {\em 2017 IEEE International Conference on Computer Vision Workshops (ICCVW)}, pages 18--26, 2017.

\bibitem{DIBENEDETTO2022117125}
Marco {Di Benedetto}, Fabio Carrara, Luca Ciampi, Fabrizio Falchi, Claudio Gennaro, and Giuseppe Amato.
\newblock An embedded toolset for human activity monitoring in critical environments.
\newblock {\em Expert Systems with Applications}, 199:117125, 2022.

\bibitem{Dukic_2023_ICCV}
Nikola Dukic, Alan Lukezic, Vitjan Zavrtanik, and Matej Kristan.
\newblock A low-shot object counting network with iterative prototype adaptation.
\newblock In {\em ICCV, Paris, France, October 1-6, 2023}, pages 18826--18835. {IEEE}, 2023.

\bibitem{10.1007/978-3-031-73247-8_18}
Michael~A. Hobley and Victor Prisacariu.
\newblock {ABC} easy as 123: {A} blind counter for exemplar-free multi-class class-agnostic counting.
\newblock In Ales Leonardis, Elisa Ricci, Stefan Roth, Olga Russakovsky, Torsten Sattler, and G{\"{u}}l Varol, editors, {\em Computer Vision - {ECCV} 2024 - 18th European Conference, Milan, Italy, September 29-October 4, 2024, Proceedings, Part {XI}}, volume 15069 of {\em Lecture Notes in Computer Science}, pages 304--319. Springer, 2024.

\bibitem{8659316}
Mohammad Hossain, Mehrdad Hosseinzadeh, Omit Chanda, and Yang Wang.
\newblock Crowd counting using scale-aware attention networks.
\newblock In {\em 2019 IEEE Winter Conference on Applications of Computer Vision (WACV)}, pages 1280--1288, 2019.

\bibitem{10.1145/3581783.3611789}
Ruixiang Jiang, Lingbo Liu, and Changwen Chen.
\newblock Clip-count: Towards text-guided zero-shot object counting.
\newblock In {\em ACM MM}, MM '23, page 4535–4545, New York, NY, USA, 2023. Association for Computing Machinery.

\bibitem{Kang_Moon_Kim_Heo_2024}
Seunggu Kang, WonJun Moon, Euiyeon Kim, and Jae-Pil Heo.
\newblock Vlcounter: Text-aware visual representation for zero-shot object counting.
\newblock {\em AAAI}, 38(3):2714--2722, Mar. 2024.

\bibitem{kirillov2023segment}
Alexander Kirillov, Eric Mintun, Nikhila Ravi, Hanzi Mao, Chloe Rolland, Laura Gustafson, Tete Xiao, Spencer Whitehead, Alexander~C. Berg, Wan-Yen Lo, Piotr Dollár, and Ross Girshick.
\newblock Segment anything, 2023.

\bibitem{DBLP:journals/corr/abs-2304-05653}
Yi Li, Hualiang Wang, Yiqun Duan, and Xiaomeng Li.
\newblock {CLIP} surgery for better explainability with enhancement in open-vocabulary tasks.
\newblock {\em CoRR}, abs/2304.05653, 2023.

\bibitem{Lin_Chan_2024}
Wei Lin and Antoni~B. Chan.
\newblock A fixed-point approach to unified prompt-based counting.
\newblock {\em AAAI}, 38(4):3468--3476, Mar. 2024.

\bibitem{Liu_2022_BMVC}
Chang Liu, Yujie Zhong, Andrew Zisserman, and Weidi Xie.
\newblock Countr: Transformer-based generalised visual counting.
\newblock In {\em BMVC}. {BMVA} Press, 2022.

\bibitem{8954153}
Weizhe Liu, Mathieu Salzmann, and Pascal Fua.
\newblock Context-aware crowd counting.
\newblock In {\em CVPR}, pages 5094--5103, 2019.

\bibitem{DBLP:journals/corr/abs-2403-05435}
Anindya Mondal, Sauradip Nag, Xiatian Zhu, and Anjan Dutta.
\newblock Omnicount: Multi-label object counting with semantic-geometric priors.
\newblock {\em CoRR}, abs/2403.05435, 2024.

\bibitem{10619063}
Vivien Patakvölgyi, Levente Kovács, and Dániel~András Drexler.
\newblock Artificial neural networks based cell counting techniques using microscopic images: A review.
\newblock In {\em 2024 IEEE 18th International Symposium on Applied Computational Intelligence and Informatics (SACI)}, pages 000327--000332, 2024.

\bibitem{Pelhan_2024_CVPR}
Jer Pelhan, Alan Luke\v{z}i?, Vitjan Zavrtanik, and Matej Kristan.
\newblock Dave - a detect-and-verify paradigm for low-shot counting.
\newblock In {\em CVPR}, pages 23293--23302, June 2024.

\bibitem{DBLP:conf/icml/RadfordKHRGASAM21}
Alec Radford, Jong~Wook Kim, Chris Hallacy, Aditya Ramesh, Gabriel Goh, Sandhini Agarwal, Girish Sastry, Amanda Askell, Pamela Mishkin, Jack Clark, Gretchen Krueger, and Ilya Sutskever.
\newblock Learning transferable visual models from natural language supervision.
\newblock In {\em Proceedings of the 38th International Conference on Machine Learning, {ICML} 2021, 18-24 July 2021, Virtual Event}, volume 139, pages 8748--8763. {PMLR}, 2021.

\bibitem{9577832}
V. Ranjan, U. Sharma, T. Nguyen, and M. Hoai.
\newblock Learning to count everything.
\newblock In {\em CVPR}, pages 3393--3402, Los Alamitos, CA, USA, jun 2021. IEEE Computer Society.

\bibitem{8099912}
Deepak~Babu Sam, Shiv Surya, and R.~Venkatesh Babu.
\newblock Switching convolutional neural network for crowd counting.
\newblock In {\em CVPR}, pages 4031--4039, 2017.

\bibitem{10483595}
Z. Shi, Y. Sun, and M. Zhang.
\newblock Training-free object counting with prompts.
\newblock In {\em 2024 IEEE/CVF Winter Conference on Applications of Computer Vision (WACV)}, pages 322--330, Los Alamitos, CA, USA, jan 2024. IEEE Computer Society.

\bibitem{8578662}
Zenglin Shi, Le Zhang, Yun Liu, Xiaofeng Cao, Yangdong Ye, Ming-Ming Cheng, and Guoyan Zheng.
\newblock Crowd counting with deep negative correlation learning.
\newblock In {\em CVPR}, pages 5382--5390, 2018.

\bibitem{Wang_Xiao_Cao_Lu_2024}
Zhicheng Wang, Liwen Xiao, Zhiguo Cao, and Hao Lu.
\newblock Vision transformer off-the-shelf: A surprising baseline for few-shot class-agnostic counting.
\newblock {\em AAAI}, 38(6):5832--5840, Mar. 2024.

\bibitem{10204688}
Jingyi Xu, Hieu Le, Vu Nguyen, Viresh Ranjan, and Dimitris Samaras.
\newblock Zero-shot object counting.
\newblock In {\em CVPR}, pages 15548--15557, 2023.

\bibitem{DBLP:journals/corr/abs-2405-11770}
Yuanwu Xu, Feifan Song, and Haofeng Zhang.
\newblock Learning spatial similarity distribution for few-shot object counting.
\newblock {\em CoRR}, abs/2405.11770, 2024.

\bibitem{10031021}
Zhiyuan You, Kai Yang, Wenhan Luo, Xin Lu, Lei Cui, and Xinyi Le.
\newblock Few-shot object counting with similarity-aware feature enhancement.
\newblock In {\em 2023 IEEE Winter Conference on Applications of Computer Vision (WACV)}, pages 6304--6313, 2023.

\bibitem{8237658}
Shanghang Zhang, Guanhang Wu, João~P. Costeira, and José M.~F. Moura.
\newblock Fcn-rlstm: Deep spatio-temporal neural networks for vehicle counting in city cameras.
\newblock In {\em ICCV}, pages 3687--3696, 2017.

\end{thebibliography}
}

\clearpage
\maketitlesupplementary
\appendix

\section{Derivation of Counting Precision and Recall}
In this section, we provide a more detailed explanation of the derivation of Eqs. 4 and 5 from the paper, specifically the formulas for calculating \textit{counting precision} and \textit{counting recall} based on the inferred quantities $c^\text{pos}$ and $c^\text{neg}$ in the context of the mosaic test. To simplify the notation, we omit the indices $i,j$ from all the involved quantities, as our focus is on a single mosaic.

As stated in the paper, we deal with counting rather than detection. Therefore, we do not know the exact nature of each inferred instance, \ie, we cannot assign a correct/incorrect label to each different detected object. However, we can still estimate the total number of true positives (TPs), false positives (FPs), and false negatives (FNs) directly from the outputs of the counting model. We make the following assumptions:

\begin{itemize}
    \item For the positive image (the top part of the mosaic), 
    \begin{equation}
    \label{eq:tp-pos}
        \text{TP}^\text{pos} =
        \begin{cases}
            c^\text{pos}, & \text{if } c^\text{pos} < \tilde{c} \\
            \tilde{c}, & \text{otherwise}
        \end{cases},
    \end{equation}
    
\noindent where $\tilde{c}$ is the ground truth of the positive class.
Indeed, if the model predicts fewer objects than the ground truth, all the predicted objects are considered correct, and the remaining ones are FNs. Conversely, if the model predicts more objects than the ground truth, only $\tilde{c}$ objects are correct, and the remaining contribute to the FPs. This situation for FNs and FPs can be directly derived from~\cref{eq:tp-pos}. In fact, given that $c^\text{pos} = \text{FP}^\text{pos} + \text{TP}^\text{pos}$, it follows that
    \begin{equation}
    \label{eq:fp-pos}
        \text{FP}^\text{pos} =
        \begin{cases}
            0, & \text{if } c^\text{pos} < \tilde{c} \\
            c^\text{pos} - \tilde{c}, & \text{otherwise}
        \end{cases}
    \end{equation}
    and provided that $\tilde{c} = \text{FN}^\text{pos} + \text{TP}^\text{pos}$, we also have
    \begin{equation}
    \label{eq:fn-pos}
        \text{FN}^\text{pos} =
        \begin{cases}
            \tilde{c} - c^\text{pos}, & \text{if } c^\text{pos} < \tilde{c} \\
            0, & \text{otherwise}
        \end{cases}
    \end{equation}

    \item For the negative image (the bottom part of the mosaic),
    the situation is simpler, given that all the contributions inferred by the model are FPs, as the TPs are identically zero, and thus also the FNs:
    \begin{align}
        \label{eq:tp-neg}
        \text{TP}^\text{neg} &= 0 \\
        \label{eq:fp-neg}
        \text{FP}^\text{neg} &= c^\text{neg} \\
        \label{eq:fn-neg}
        \text{FN}^\text{neg} &= 0
    \end{align}
\end{itemize}

With these quantities defined, we can introduce the \textit{counting precision} and the \textit{counting recall}, starting from their definitions in terms of TPs, FPs, and FNs.

\subsection{Counting Precision}
We start with the definition of precision, which is the following:
\begin{equation}
    P = \frac{\text{TP}}{\text{TP} + \text{FP}}
\end{equation}
Considering that the TPs and FPs are the sums of the respective contributions from the positive and negative parts of the mosaic -- \ie, $\text{TP} = \text{TP}^\text{pos} + \text{TP}^\text{neg}$ and $\text{FP} = \text{FP}^\text{pos} + \text{FP}^\text{neg}$ -- we obtain the precision expressed in terms of the quantities computed in~\cref{eq:tp-pos,eq:fp-pos,eq:fn-pos,eq:fp-neg}.
Substituting and simplifying, we obtain:
\begin{equation}
\label{eq:cntp}
    P = 
    \begin{dcases}
        \frac{c^\text{pos}}{c^\text{pos} + c^\text{neg}}, & \text{if } c^\text{pos} < \tilde{c} \\
        \frac{\tilde{c}}{c^\text{pos} + c^\text{neg}}, & \text{otherwise}
    \end{dcases}
\end{equation}
which we can rewrite in a simpler manner as: 
\begin{equation}
    P = \frac{\min(c^\text{pos},\tilde{c})}{c^\text{pos} + c^\text{neg}}.
\end{equation}
This quantity is averaged among all the possible mosaics, which are $N(N-1)$ (for each image, there are $N-1$ possible mosaics), to obtain the final formula for the counting precision reported in the paper.

\subsection{Counting Recall}
The same idea used for deriving the counting precision can also be employed to compute the counting recall.
The recall is defined as:
\begin{equation}
    R = \frac{\text{TP}}{\text{TP} + \text{FN}}
\end{equation}
Even in this case, TPs and FNs are the sums of the respective contributions from the positive and negative parts of the mosaic -- \ie, $\text{TP} = \text{TP}^\text{pos} + \text{TP}^\text{neg}$ and $\text{FN} = \text{FN}^\text{pos} + \text{FN}^\text{neg}$. We obtain the precision expressed in terms of quantities computed in \cref{eq:tp-pos,eq:fn-pos,eq:tp-neg,eq:fn-neg}.
Substituting and simplifying, we obtain:
\begin{equation}
\label{eq:cntr}
    R = 
    \begin{dcases}
        \frac{c^\text{pos}}{\tilde{c}}, & \text{if } c^\text{pos} < \tilde{c} \\
        1, & \text{otherwise}
    \end{dcases}
\end{equation}
which we can rewrite as:
\begin{equation}
    R = \frac{\min(c^\text{pos},\tilde{c})}{\tilde{c}}.
\end{equation}

\noindent Again, this quantity is averaged in the same way as counting precision to obtain the final formula reported in the paper.

\section{Derivation of Normalized Mean of Negative predictions (NMN)}
NMN, as reported in the paper, is the absolute counting error computed by prompting the model with the negative classes normalized by the ground truth of the positive class.
Formally, the main involved quantity computed for each image $I_i$ prompted with the negative class $P_j$ is given by:
\begin{equation}
    n_{ij} = \frac{|c_{ij} - \tilde{c}_{ij}^\text{neg}|}{\tilde{c}_i}, \quad i \neq j
\end{equation}
where $\tilde{c}_{ij}^\text{neg}$ is the ground truth corresponding to the image prompted with the negative class, which is identically zero for $i \neq j$.
Therefore, the numerator simplifies from $|c_{ij} - \tilde{c}_{ij}^\text{neg}|$ to $c_{ij}$ (we assume the count predicted by the model is always positive).
All the $N_{ij}$ are then averaged over all the $N$ images, each one prompted with all the possible $N-1$ negative prompts:
\begin{align}
    \text{NMN} &= \frac{1}{N}\sum_{i=1}^N \frac{1}{N-1}\sum_{\substack{j=1 \\ j \neq i}}^{N} n_{ij} \\
    &= \frac{1}{N}\sum_{i=1}^N \frac{1}{N-1}\sum_{\substack{j=1 \\ j \neq i}}^{N} \frac{c_{ij}}{\tilde{c}_i} \\
    &= \frac{1}{N(N-1)}\sum_{i=1}^N \frac{1}{\tilde{c}_i}\sum_{\substack{j=1 \\ j \neq i}}^{N} c_{ij}
\end{align}
which is the Eq. 2 reported in the paper.

\begin{table*}[t]
    \centering
    \caption{We report the results for \textbf{DAVE} on the \textbf{test set} of FSC147, varying the clustering threshold $\tau$ (lowering it from the original 0.17 to 0.10, and modifying the inference procedure (\textit{Mod. Inf.} column) obtained by feeding the model also with the reference positive class.}%
    \label{tab:dave-ablation}%
    \newcolumntype{L}{>{\arraybackslash}m{.25\linewidth}}%
    \newcolumntype{C}{>{\centering\arraybackslash}X}
    %\tiny%
    \setlength{\tabcolsep}{1pt}
    \begin{tabularx}{0.95\linewidth}{CC|CC|CCC|CC}
        \toprule
        & &  \multicolumn{2}{c|}{Negative Test} & \multicolumn{3}{c|}{Mosaic Test} & \multicolumn{2}{c}{Classic} \\
        \cmidrule(lr){3-4} \cmidrule(lr){5-7} \cmidrule(lr){8-9}
        \centering $\tau$ & Mod. Inf. & NMN $\downarrow$ & PCCN $\uparrow$ & CntP $\uparrow$ & CntR $\uparrow$ & CntF1 $\uparrow$ & MAE $\downarrow$ & RMSE $\downarrow$ \\
        \midrule
        \midrule
        0.17 & \xmark & 1.05 & 37.02 & 0.686 & 0.811 & 0.700 & 15.16 & 103.49 \\
        0.10 & \xmark & 1.05 & 37.02 & 0.743 & 0.805 & 0.732 & 15.16 & 103.49 \\

        0.17 & \cmark & 0.08 & 97.45 & 0.831 & 0.803 & 0.784 & 15.11 & 103.48 \\
        0.10 & \cmark & 0.08 & 97.62 & 0.843 & 0.799 & 0.790 & 15.23 & 103.53 \\
        \bottomrule
    \end{tabularx}
\end{table*}

%%%%%%%%% BODY TEXT
\section{DAVE Inference Details}
We performed small changes to the inference code to prepare the DAVE model for our benchmark. This small update drastically improved DAVE on \benchmarkacron{}, unblocking its full potential.

Indeed, although DAVE has been designed to be resilient to images with multiple classes, the method assumes that it is prompted by one of the classes that are surely present in the image. %This obviously is not the case for the negative test in \benchmarkacron{}. 
In these cases, the model just considers the object class whose CLIP embedding is more similar to the provided prompt instead of allowing for zero matches based on a certain score threshold. % which is exactly the case we are not interested in. This is again possibly due to the fact that either the method is presented with images having multiple objects  
If DAVE is prompted with a class not present in the one-class-only image, the original implementation ignores the CLIP-based proposal filtering. The outcome is catastrophic, especially for our \textit{negative test}, as DAVE outputs the same count regardless of the input text prompt. For this reason, we modified DAVE to filter the proposals associated with the sole present cluster based on the input text. To compute the threshold to decide if the cluster proposals match the provided caption, we also fed the model with the positive class to have a CLIP upper-bound score as a reference. As in the original implementation, the proposals are kept if their CLIP score is higher than 85\% of this reference CLIP score. Notice that this inference procedure would be difficult in real scenarios in which the positive class is not known a-priori. However, since the positive class can be obtained through image classification -- and image classification is a well-established and solved problem in computer vision -- we assume that, in real use-case scenarios, it is possible to derive a reliable positive class label using state-of-the-art image classifiers.

We also noticed that the outcome on our benchmark is very dependent on the clustering threshold $\tau$ used during the spectral clustering phase. Particularly, we observed that the original $\tau=0.17$ was too high to correctly detect the two clusters corresponding to the two images in the mosaics. For this reason, in the main paper experiments, we set $\tau=0.10$.

In~\cref{tab:dave-ablation}, we report an ablation study about the model's behavior (i) with and without modification to the inference strategy, and (ii) the original and changed $\tau$ parameter.
As we can notice, the clustering threshold does not affect the negative test, where only one object cluster is always found. Our modification, which injects positive classes as a reference, originates a strong model from the negative test perspective, with an NMN of only 0.08. 
Concerning the mosaic model, the lowering of the $tau$ threshold, together with the improved inference procedure, helps raise the counting precision and, in turn, the counting F1-score by more than 12\% with respect to the original implementation.

It is interesting to notice how these hyper-parameters have no effect on the class-specific classic counting metrics (MAE and RMSE), again proving the need for benchmarks like \benchmarkacron{} to effectively evaluate prompt-based counting models.

\section{TFPOC Density Maps Creation}
TFPOC is a detection-based method that localizes objects to count using the powerful SAM model~\cite{kirillov2023segment}. 
%Therefore, it is a detection-based model that counts the number of output detections. 
For this reason, it never really computes a density map, which is the main output interface used to prepare the predictions for the mosaic test and produce the qualitative visualization. To prepare the density maps, we simply plotted the region centers as small dots, each having an area of 1 (as is usually done for preparing ground truth density maps from dot annotations).

\section{More Qualitative Results}
In~\cref{fig:qualitative-results-negative-1}, we present four images provided as input to the model, each paired with different negative classes. Notably, all methods except DAVE count the negative classes, often predicting a number of instances comparable to -- or even exceeding -- the ground truth for the positive class. In contrast, DAVE consistently predicts zero instances, demonstrating the effectiveness of the proposed inference modification.

In~\cref{fig:qualitative-results-mosaic-1}, we present additional results from the mosaic test, illustrating how the models often struggle to count exclusively the correct class. Notably, while DAVE demonstrates strong performance in distinguishing the sole positive class from negative ones and achieves impressive results on the \benchmarkacron{} metrics for the mosaic test, it occasionally suffers catastrophic failures, incorrectly swapping the positive class with a negative one.

% \begin{figure*}[t]
%     \centering
%     \begin{subfigure}{\linewidth}
%         \includegraphics[width=.90\linewidth,trim={1cm 1cm 0 0},clip]{imgs/suppl/qualitative_mosaics_1.pdf}
%         \caption{}
%         \label{fig:subfigA}
%     \end{subfigure}
%     \begin{subfigure}{\linewidth}
%         \includegraphics[width=.90\linewidth,trim={1cm 1cm 0 0},clip]{imgs/suppl/qualitative_mosaics_2.pdf}
%         \caption{}
%         \label{fig:subfigB}
%     \end{subfigure}
%     \caption{This figure shows, for each model, the output density maps for three different \textit{(mosaic, input prompt)} pairs. In each figure, the count reported in the blue box is $c^\text{pos}_{ij}$, while the count reported in the red box corresponds to $c^\text{neg}_{ij}$. %We can notice how the models incorrectly attend to the instances from the negative image in the mosaic, although most of them provide a good estimate for the positive instances in the upper image of the mosaic.}
%     }
%     \label{fig:qualitative-results-mosaic}
% \end{figure*}

\begin{figure*}[t]
    \centering
    \begin{subfigure}{\linewidth}
        \includegraphics[width=.9\linewidth,trim={1cm 1cm 0 0},clip]{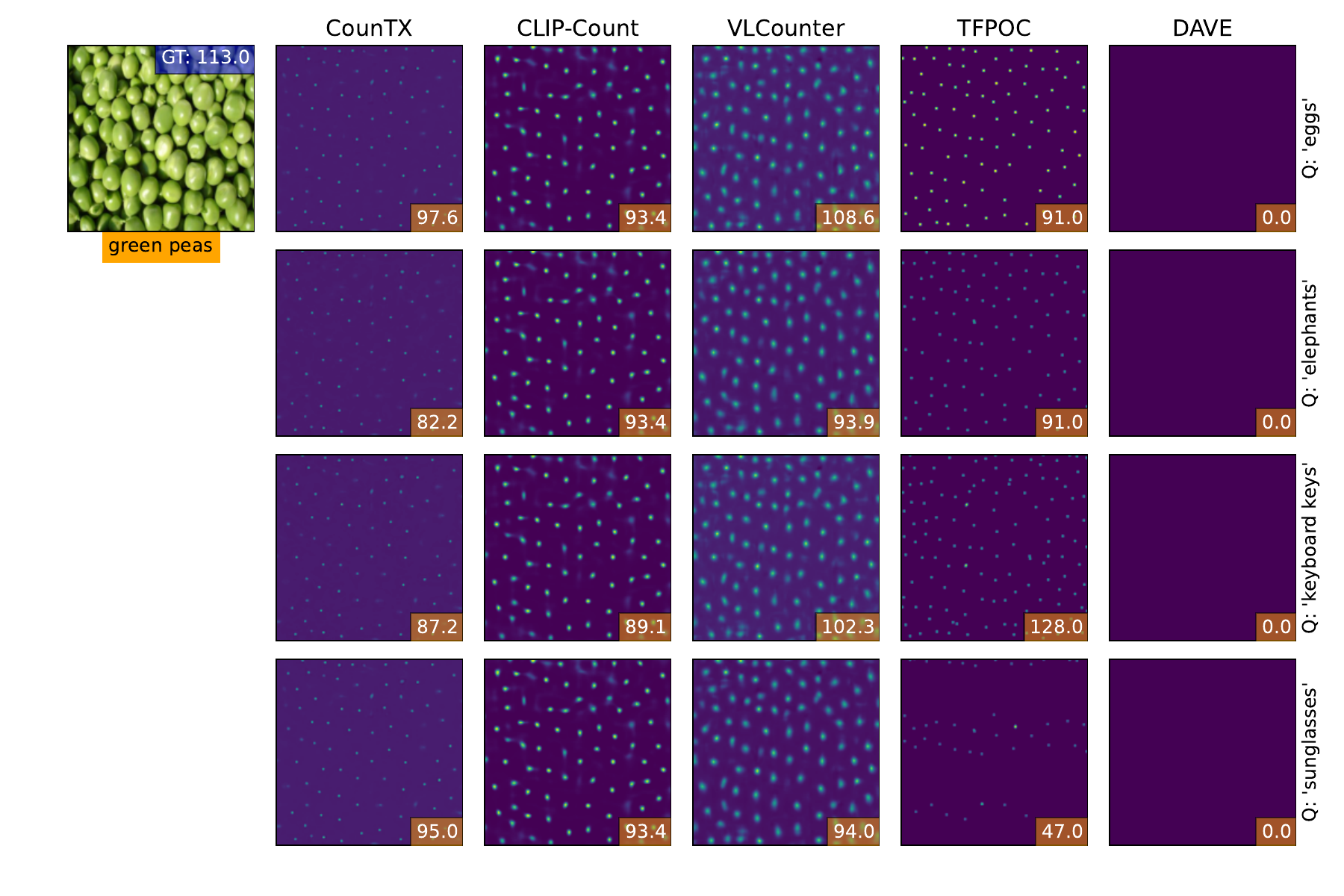}
        \caption{}
        \label{fig:subfigA}
    \end{subfigure}
    \begin{subfigure}{\linewidth}
        \includegraphics[width=.9\linewidth,trim={1cm 1cm 0 0},clip]{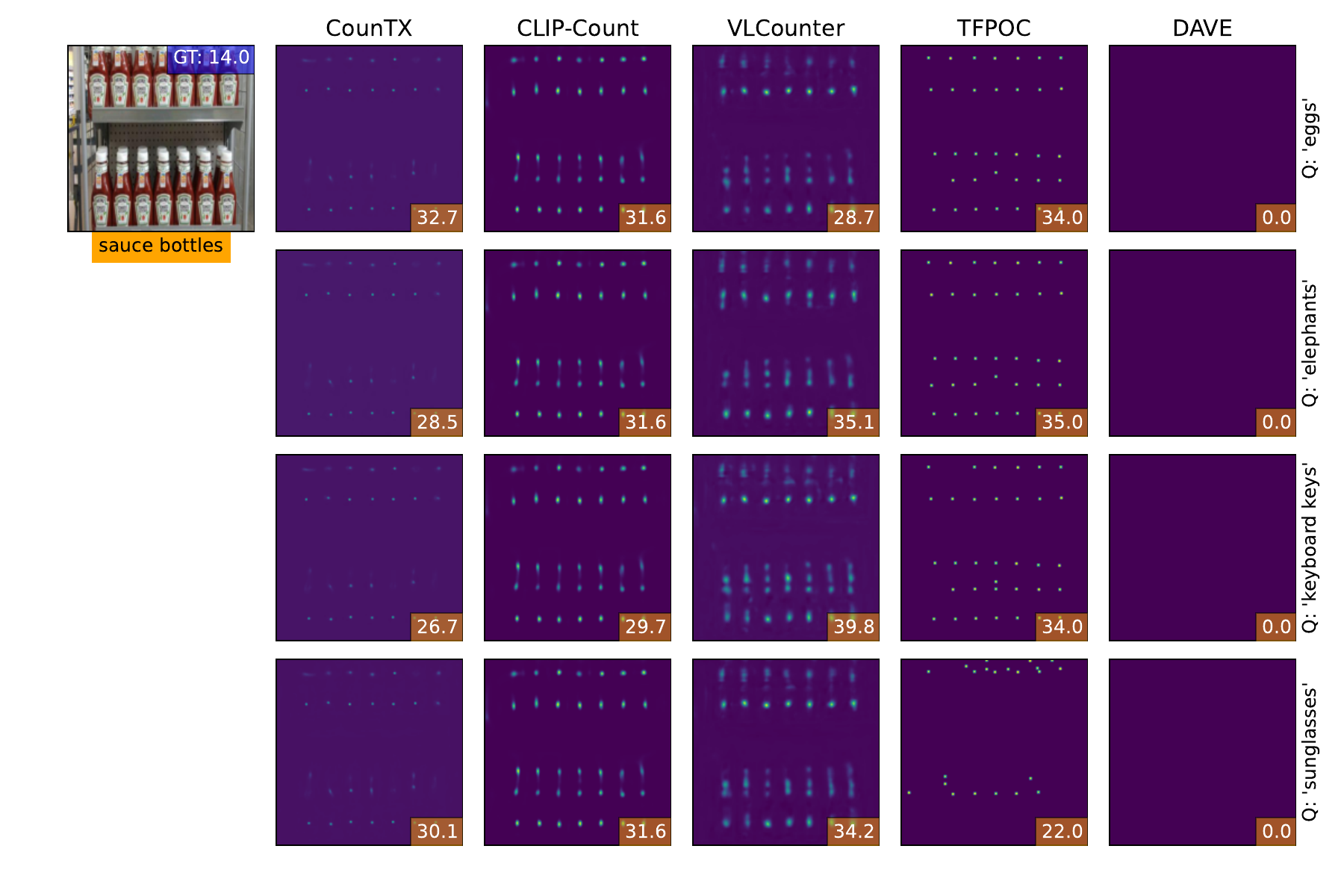}
        \caption{}
        \label{fig:subfigB}
    \end{subfigure}
    \caption{For each model, we report the density maps obtained when probing them with four different negative classes (\textit{eggs, elephants, keyboard keys, sunglasses}) reported in the right-hand side of each row.}
    \label{fig:qualitative-results-negative-1}
    
\end{figure*}
\begin{figure*}[t]\ContinuedFloat
    \centering
    \begin{subfigure}{\linewidth}
        \includegraphics[width=.9\linewidth,trim={1cm 1cm 0 0},clip]{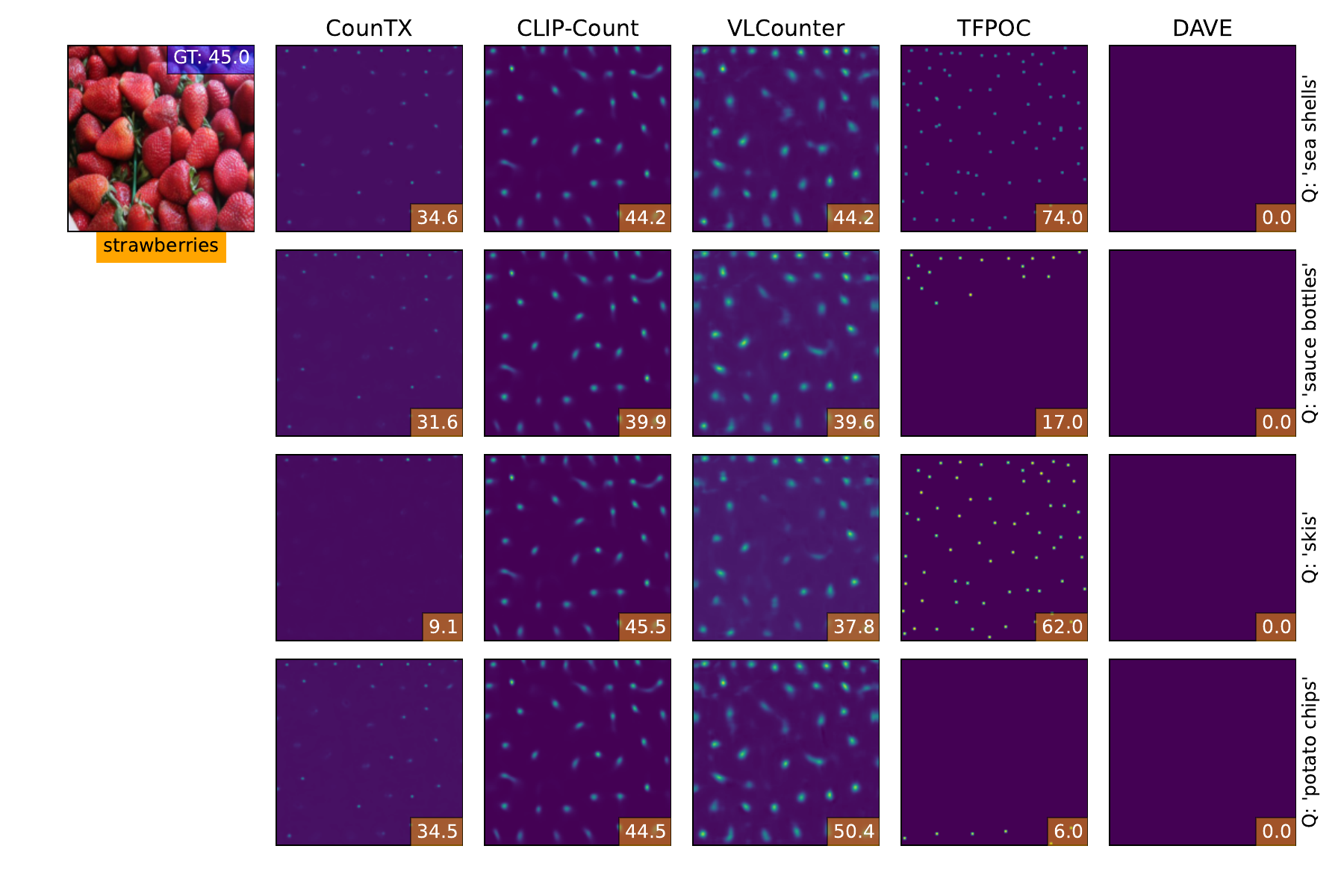}
        \caption{}
        \label{fig:subfigC}
    \end{subfigure}
    \begin{subfigure}{\linewidth}
        \includegraphics[width=.9\linewidth,trim={1cm 1cm 0 0},clip]{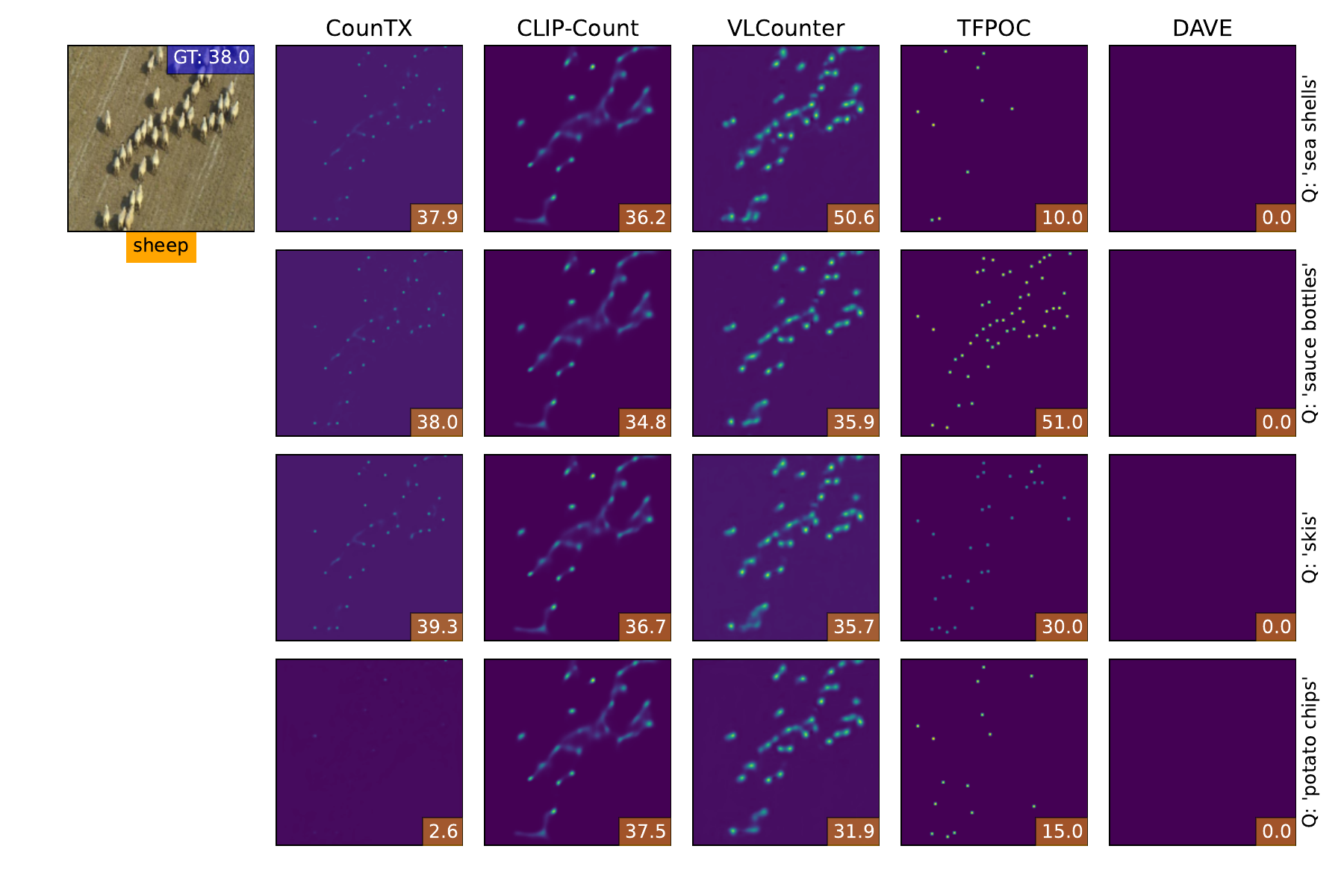}
        \caption{}
        \label{fig:subfigD}
    \end{subfigure}
    \caption{For each model, we report the density maps obtained when probing them with four different negative classes (\textit{sea shells, sauce bottles, skis, potato chips}) reported in the right-hand side of each row (cont).}
    \label{fig:qualitative-results-negative-2}
\end{figure*}

\begin{figure*}[t]
    \centering
    \includegraphics[width=.90\linewidth,trim={1cm 1cm 0 0},clip]{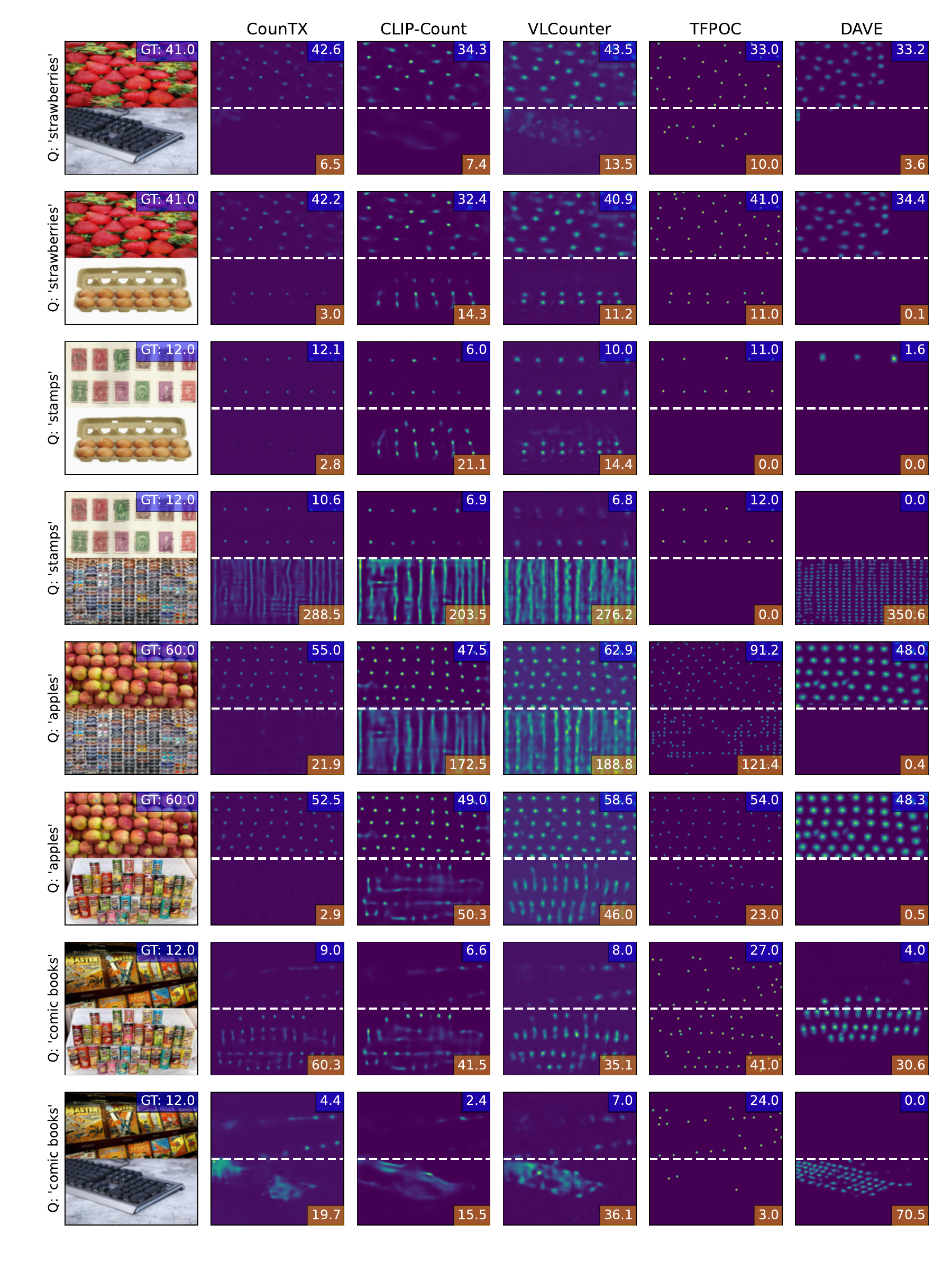}
    
    \caption{For each model, we report the output density maps for three different \textit{(mosaic, input prompt)} pairs. In each figure, the count reported in the blue box is $c^\text{pos}_{ij}$, while the count reported in the red box corresponds to $c^\text{neg}_{ij}$. %We can notice how the models incorrectly attend to the instances from the negative image in the mosaic, although most of them provide a good estimate for the positive instances in the upper image of the mosaic.}
    }
    \label{fig:qualitative-results-mosaic-1}
\end{figure*}
\begin{figure*}[t]\ContinuedFloat
    \centering
    \includegraphics[width=.90\linewidth,trim={1cm 1cm 0 0},clip]{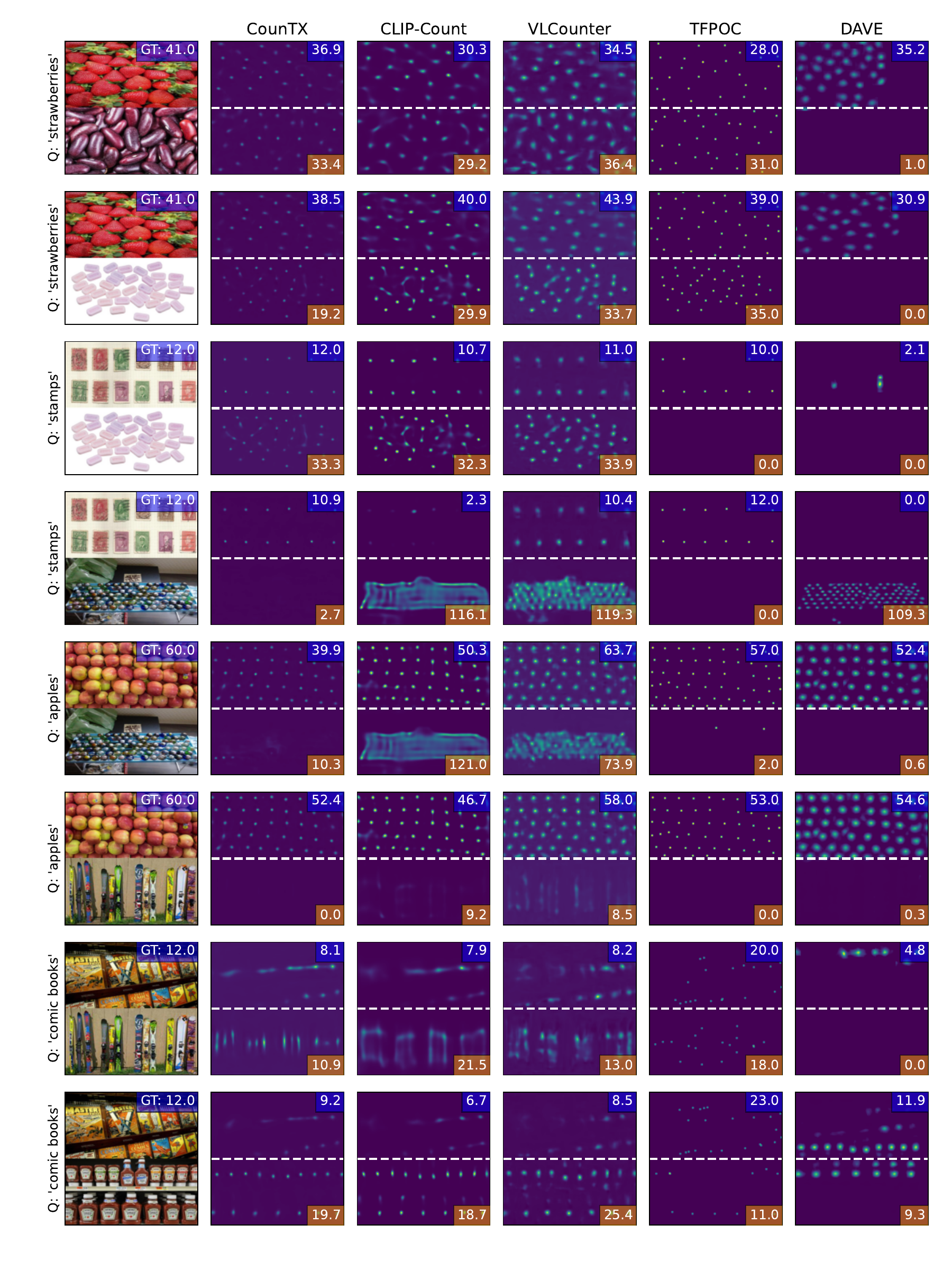}
    
    \caption{For each model, we report the output density maps for three different \textit{(mosaic, input prompt)} pairs. In each figure, the count reported in the blue box is $c^\text{pos}_{ij}$, while the count reported in the red box corresponds to $c^\text{neg}_{ij}$ (cont). %We can notice how the models incorrectly attend to the instances from the negative image in the mosaic, although most of them provide a good estimate for the positive instances in the upper image of the mosaic.}
    }
    \label{fig:qualitative-results-mosaic-2}
\end{figure*}

\end{document}